\documentclass[11pt,a4paper]{article}

\usepackage[T1]{fontenc}
\usepackage[utf8]{inputenc}
\usepackage[margin=0.95in]{geometry}
\usepackage{lmodern}
\usepackage{microtype}
\usepackage{setspace}
\usepackage{amsmath,amssymb,amsfonts,mathtools}
\usepackage{bm}
\usepackage{booktabs}
\usepackage{multirow}
\usepackage{array}
\usepackage{adjustbox}
\usepackage{enumitem}
\usepackage{xcolor}

\usepackage[numbers,sort&compress,square]{natbib}
\definecolor{linkblue}{RGB}{20,70,150}
\definecolor{citegreen}{RGB}{20,110,70}

\usepackage[
  colorlinks=true,
  linkcolor=linkblue,
  citecolor=citegreen,
  urlcolor=linkblue
]{hyperref}
\usepackage[nameinlink,noabbrev]{cleveref}

\newcommand{\DiD}{\mathrm{DiD}}
\newcommand{\Cost}{\mathrm{Cost}}
\newcommand{\code}[1]{\texttt{#1}}

\title{\textbf{When Nothing Announces the Change:\\Derivation-Certificate Memory under Silent Environment Drift}}

\author{
  \textbf{Dawei Ren} \qquad \textbf{Yan Zhang}\\[4pt]
  National Key Laboratory of Heterogeneous Integrated Microsystems, SIMIT,\\
  Chinese Academy of Sciences, Shanghai, 200050, China\\[4pt]
  \texttt{\href{mailto:rendawei23@mails.ucas.edu.cn}{rendawei23@mails.ucas.edu.cn}}, \quad
  \texttt{\href{mailto:zhangyan232@mails.ucas.ac.cn}{zhangyan232@mails.ucas.ac.cn}}
}

\date{}

\begin{document}
\maketitle

\begin{abstract}
\noindent
Long-lived tool-using language agents rely on cross-episode memory to amortize expensive multi-step exploration, yet external application states mutate between sessions. Prior work has established that agent memories go stale, that similarity retrieval can rank outdated facts above current ones, and that provenance graphs can propagate invalidation when new evidence arrives. Across this literature, however, maintenance relies on an explicit \emph{change trigger}: a new document, a version bump, a typed drift event, or a user correction. We study \textbf{silent environment drift}, where underlying databases and APIs change between episodes with no external signal. To isolate this regime, we introduce \textbf{StaleWorld}, a benchmark of $787$ executable AppWorld tasks across $3{,}608$ materialized worlds featuring paired \emph{stable} and \emph{drift} arms, audited current-world ground truth, answer-preserving background noise, coupled global drift, and template-disjoint held-out tasks; the paired stable control enables per-task difference-in-differences ($\DiD$) that separates drift-induced memory failure from intrinsic task difficulty. We then propose \textbf{Derivation-Certificate Memory (DCM)}, an epistemic memory architecture that treats stored notes not as timeless answers but as \emph{warranted beliefs} backed by executable read-only premise certificates verified against the live environment at read time. We instantiate three complementary variants: \textbf{DCM-Core} (minimal operator-level boundary, join, and target certificates), \textbf{DCM-Auto} (zero-prior LLM-synthesized read-only verifiers constructed opportunistically from remaining episode budget under sandbox write-interception and solve-cost gates), and \textbf{DCM-Repair} (witness-guided belief revision via live premise diffs plus surface-anchor-gated verified action trajectory replay). On StaleWorld's $700$ covered tasks, standard retrieval memory suffers a catastrophic \textbf{65.0-point} post-drift collapse ($0.834 \to 0.184$ in post-drift episodes, $\DiD = -0.648$), submitting stale answers on \textbf{100.0\%} of drifted answer episodes where prior memory existed and committing \textbf{143} irreversible wrong-target actions ($16.8\%$) versus $1$ ($0.1\%$) for memoryless execution. \textbf{DCM-Core} limits post-drift loss to \textbf{10.0 points} ($0.807 \to 0.706$, surpassing memoryless re-solving's $0.689$ at \textbf{25.7\% lower cost per success}), rejects \textbf{100.0\%} ($537/537$) of invalidated certificates at drift onset, records \textbf{zero stale acceptances} across $4{,}615$ verified cache hits, and reduces post-drift wrong-target actions by \textbf{143$\times$} ($1$ vs.\ $143$). Across $678$ answer-preserving background-noise tasks, \textbf{DCM-Core} triggers \textbf{$0 / 3{,}271$ ($0.00\%$)} false suspensions; under coupled global drift, it maintains $0.708$ accuracy with \textbf{$0 / 822$ ($0.0\%$)} wrong-target actions. Extending beyond operator builders, \textbf{DCM-Repair} reaches \textbf{$0.723$} post-drift accuracy on a stratified $11$-policy comparison (ranking first overall) and \textbf{$0.891$} on $10$-episode stochastic multi-shock streams (beating parameterized Time-To-Live baselines by \textbf{+19.0 points} at \textbf{40\%--48\% lower cost}); \textbf{DCM-Auto} reaches \textbf{$0.693$} post-drift accuracy ($0.853$ on answer tasks and $0.533$ on action tasks, with $0$ wrong-target actions and $0$ stale submissions, surpassing memoryless execution at \textbf{23.7\% lower cost}) and generalizes zero-prior across $29$ unseen held-out templates ($0.559$ post-drift and $0.651$ stationary accuracy, beating retrieval by \textbf{+19.1 points} and memoryless re-solving at \textbf{23.1\% lower cost}); and all findings replicate across paraphrased zero-key shared semantic pools ($+50.0$ to $+54.2$ points over retrieval, $0$--$1$ vs.\ $73$ wrong-target actions), multiple random seeds ($\mathrm{SD} = 0.015$), and a $3$B-active Sparse-MoE architecture ($4.2\times$ retrieval accuracy). Together, StaleWorld and DCM establish that when nothing announces a world change, reliable agent memory requires priced, premise-level live verification.
\end{abstract}

\section{Introduction}
\label{sec:intro}

Consider a personal assistant agent instructed to reimburse a friend for dinner on Venmo using a phone number provided by the user. In Episode~1, the agent looks up the number in the contacts app, resolves the friend's Venmo email, sends the payment, and stores the resolved target in memory. Before Episode~4, the phone number is reassigned to a different contact in the database---with no notification, system log, or user message. When asked again, an agent using standard retrieval memory recalls a semantically identical task note, skips contact resolution, and transfers money to a stranger.

Such failures are not hallucinations, nor are they retrieval errors. The stored note is faithful to past observations and matches the user's query with maximal semantic similarity. The failure is \emph{epistemic}: \textbf{semantic similarity does not imply current epistemic validity}. While cross-episode memory makes tool-using agents dramatically faster and more consistent in stationary environments, unverified memory turns past competence into confident, high-consequence error when the environment mutates. Furthermore, the pragmatic stakes are sharply asymmetric between \emph{answering} and \emph{acting}: a stale question-answering memory yields an outdated string, whereas a stale action memory executes irreversible side effects---misdirected payments, spurious messages, or destructive library deletions---in external systems.

Recent research has elevated staleness to a central challenge in agent memory. Benchmarks and analyses in conversational memory demonstrate that models struggle to update beliefs under implicit conflict~\citep[STALE;][]{chao2026stale}, that similarity search favors entrenched outdated facts~\citep[StaleBench;][]{murshed2026stalebench}, that outdated procedural memory can underperform memoryless execution~\citep[WK-Drift;][]{banerjee2026wkdrift}, and that maintenance rather than storage capacity is the primary bottleneck~\citep[Supersede;][]{patel2026supersede}. Concurrently, structured maintenance systems---including PACMI~\citep{wang2026pacmi}, PALIMPSEST~\citep{beltran2026palimpsest}, MemTX~\citep{li2026memtx}, DASR~\citep{wang2026dasr}, TRACE~\citep{xiong2026trace}, and VAIL~\citep{wang2026vail}---introduce provenance graphs, transactional clocks, and executable record gates. These advances decisively refute the naive strawman that existing memory systems ignore world change.

Yet existing maintenance mechanisms share a foundational structural assumption: \textbf{change announces itself}---or else every record is checked unconditionally regardless of cost. Write-time truth-maintenance systems (PACMI, PALIMPSEST, CUPMem) cascade invalidations when a new evidence item, retraction, or explicit correction enters the memory stream. Event-driven benchmarks (WK-Drift) supply typed drift event tuples directly to the agent. Transactional and multi-agent workspace verifiers (MemTX, DASR, TRACE) check consistency against observed commits or return deltas. In deployed tool environments---databases, financial ledgers, CRMs, inbox states, and media libraries---external state routinely mutates \emph{silently} between agent invocations. Under \textbf{silent environment drift}, write-time invalidation never fires because no write arrives, and internal belief-graph consistency checks pass vacuously because the external mutation was never ingested into the graph. Consequently, epistemic validity can only be established \textbf{at read time by probing the live environment}, and because live environment queries and LLM reasoning both carry non-zero cost, maintenance becomes an explicit economic decision among trusting, verifying, and re-solving.

We address this gap through a paired benchmark and a family of epistemic memory mechanisms:

\begin{enumerate}[leftmargin=1.6em,itemsep=0.25em]
  \item \textbf{Isolating Silent Drift with Paired Causal Controls (StaleWorld).} We construct \textbf{StaleWorld}, a benchmark of $787$ executable multi-app tasks built on AppWorld~0.2.0~\citep{trivedi2024appworld} across $3{,}608$ materialized database worlds, spanning paired \emph{stable} and \emph{local-drift} arms, two \emph{answer-preserving noise} arms, a coupled \emph{global-drift} arm, $200$ template-disjoint transfer tasks, and $87$ free-form held-out tasks across $29$ unseen official templates. Because upstream AppWorld ground truth becomes obsolete after database mutations, StaleWorld recomputes and audits current-world ground truth across all $3{,}608$ worlds. Crucially, running identical instructions, budgets, and models on paired stable and drift worlds enables per-task difference-in-differences ($\DiD$) estimation, cleanly isolating drift-specific memory loss from task difficulty and cold-start variance.
  \item \textbf{Warranted Belief Memory via Live Derivation Certificates (DCM).} Drawing on classic justificationist epistemology and truth maintenance~\citep{doyle1979truth}, we formulate \textbf{Derivation-Certificate Memory (DCM)}: a memory note is reusable only while the minimal environmental premises that \emph{warranted} its answer or action target still hold in the live world. We formalize the soundness--cost trade-off between \emph{Exact Dependency Certificates} and lightweight \emph{$k$-Horizon Boundary Certificates}, pair live verification with a read-time \emph{derivation guard} against poisoned or self-contradictory notes, and price verification against re-solving.
  \item \textbf{Bridging Minimal Theory, Zero-Prior Generalization, and Action Recovery.} To ensure that derivation certificates are neither restricted to hand-engineered templates nor wasteful when verification fails, we unify three progressive mechanisms: \textbf{DCM-Core} (minimal operator-level certificates characterizing theoretical probe efficiency and zero noise false-suspension), \textbf{DCM-Auto} (opportunistically in-budget LLM-synthesized read-only Python verifiers validated via static/snapshot write interception and measured solve-cost gates, requiring zero human builders on unseen templates), and \textbf{DCM-Repair} (witness-guided belief revision that injects live premise diffs upon invalidation, paired with surface-anchor-gated verified action trajectory replay and contextual fallback).
  \item \textbf{Comprehensive Empirical Characterization.} Across $149{,}400$ evaluated episodes with zero infrastructure errors, we evaluate lifecycle accuracy, post-drift $\DiD$, conditional stale submissions, irreversible wrong-target actions, background-noise false suspensions, adversarial memory poisoning, out-of-horizon rank jumps, paraphrased cross-task semantic memory pools with zero task-identifier keying, stochastic multi-shock streams against parameterized Time-To-Live (TTL) baselines, cost-weight sensitivity ($\alpha \in [0, 100]$), multi-seed stability, and Dense-27B versus Sparse-MoE-35B model architectures.
\end{enumerate}

\section{Related Work}
\label{sec:related}

\subsection{Agent Memory Architectures and Retrieval}
Contemporary agent memory systems---including MemGPT/Letta~\citep{packer2023memgpt}, A-MEM~\citep{xu2025amem}, Mem0~\citep{chhikara2025mem0}, and Graphiti/Zep~\citep{rasmussen2025zep}---organize interaction history into hierarchical context buffers, linked Zettelkasten notes, or temporally indexed knowledge graphs. Many incorporate recency decay or write-time supersession when conflicting statements are extracted. The limitation we target is not an absence of update operators, but \emph{trigger blindness}: their update machinery is driven by incoming observations. When external application tables change out-of-band without an incoming message, similarity retrieval faithfully surfaces high-matching notes whose underlying premises have silently expired.

\subsection{Conversational Staleness and Drift Benchmarks}
STALE~\citep{chao2026stale} demonstrates that LLMs fail to revise downstream behavior under implicit conversational conflict and proposes write-time consolidation (CUPMem). Temporal Belief Reconstruction~\citep{hirota2026tbr} walks declared \code{derived\_from} links over bitemporal question-answering dialogues, while Supersede~\citep{patel2026supersede} highlights the memory-update gap in user-fact streams. Closer to tool use, \textbf{WK-Drift}~\citep{banerjee2026wkdrift} pairs Era-1 and Era-2 tasks with post-drift gold and shows that stale procedural memory can hurt performance; however, WK-Drift explicitly scopes out drift detection by providing typed drift event tuples directly to the agent, and lacks a difficulty-matched stable control arm or executable database side effects. \textbf{StaleBench}~\citep{murshed2026stalebench} studies stale fact retrieval and storage pricing over chronological text streams without executable multi-app procedures or actions. StaleWorld advances this line along three axes: \emph{unannounced} database mutations, a \emph{paired stable control arm} for per-task $\DiD$, and \emph{executable action-target evaluation} where stale reuse causes tangible external mutations.

\subsection{Truth Maintenance, Provenance, and Use-Time Verification}
Dependency-directed belief revision traces back to Truth Maintenance Systems~\citep{doyle1979truth}, AGM theory~\citep{alchourron1985logic}, and database incremental view maintenance~\citep{gupta1995maintenance}. Recent agent frameworks adapt this lineage: \textbf{PACMI}~\citep{wang2026pacmi} cascades validity states across a provenance graph upon new evidence arrival; \textbf{PALIMPSEST}~\citep{beltran2026palimpsest} retracts downstream beliefs when source reports are withdrawn; \textbf{MemTX}~\citep{li2026memtx} enforces transactional commits over shared memory; and \textbf{DASR}~\citep{wang2026dasr} flags \code{Needs-Recompute} states from observed state transitions. Because silent drift provides no arrival event, internal provenance cascades remain dormant.

The closest use-time verifiers are \textbf{VAIL}~\citep{wang2026vail}, which attaches an executable live-environment check to each memory record under the principle that \emph{retrieval authorizes inspection, not use}, and \textbf{TRACE}~\citep{xiong2026trace}, which gates multi-agent role re-entry via grounded premise verification under a token budget. DCM shares the separation of retrieval from epistemic warrant, but differs in \emph{certified unit}, \emph{economics}, and \emph{revision} (\Cref{tab:related-work}): rather than binary record-level admission with mandatory checking, DCM certifies the minimal derivation premises of a specific answer or action target, prices verification against re-solving ($C_{\text{verify}}$ vs.\ $C_{\text{solve}}$), synthesizes self-checked verifiers on unseen tasks (\textbf{DCM-Auto}), and converts failed verifications into structured counterfactual diagnostics (\textbf{DCM-Repair}).

\subsection{Certificates, Caching Economics, and Memory Poisoning}
Recent systems invoke ``certificates'' in distinct senses: \textbf{FinCacheServe}~\citep{zeng2026fincacheserve} attaches document version fingerprints to materialized RAG answers; \textbf{Many Steps into One}~\citep{zhou2026manysteps} constructs statistical e-process certificates over audited outcomes; and \textbf{Reclaim}~\citep{kwon2026reclaim} stores write-time recomputation bases to repair lossy compression. None probe live external databases when no version store signals a mutation. Separately, adversarial memory poisoning studies---PoisonedRAG~\citep{zou2024poisonedrag}, AgentPoison~\citep{chen2024agentpoison}, A-MemGuard~\citep{wei2025amemguard}, and DriftGuard~\citep{pan2026driftguard}---address notes corrupted at write time or hijacking execution trajectories. While provenance systems typically assume faithful writes, DCM's read-time \emph{derivation guard} verifies that a stored answer or target is logically entailed by its own certificate before probing the world, neutralizing injected notes that contradict valid premises.

\begin{table}[t]
\centering
\caption{Conceptual comparison of StaleWorld and DCM against representative agent memory maintenance benchmarks and frameworks.}
\label{tab:related-work}
\begin{adjustbox}{max width=\textwidth}
\small
\begin{tabular}{llllllll}
\toprule
\textbf{System / Benchmark} & \textbf{Substrate} & \textbf{Change Trigger} & \textbf{Paired Stable Arm} & \textbf{Current-World Gold} & \textbf{Executable Actions} & \textbf{Certified Unit} & \textbf{Priced Gate} \\
\midrule
PACMI~\citep{wang2026pacmi} & Text memory graph & New evidence arrival & No & No & No & Record & No \\
PALIMPSEST~\citep{beltran2026palimpsest} & Report stream & Explicit retraction & No & Partial & No & Derived belief & Partial \\
MemTX / DASR~\citep{li2026memtx,wang2026dasr} & Shared state graph & Observed commit / transition & No & No & Action gate & Transaction / record & No \\
TRACE~\citep{xiong2026trace} & Multi-agent workspace & Return delta & Return eps & Internal state & Role admission & Record view & Token budget \\
VAIL~\citep{wang2026vail} & Tool/agent memory & Unconditional on use & No & No & Tool-call gate & Record & No \\
FinCacheServe~\citep{zeng2026fincacheserve} & Versioned RAG docs & Version bump & No & No & No & Doc fingerprint & Binary \\
WK-Drift~\citep{banerjee2026wkdrift} & QA / code / clinical & Supplied event tuple & No & Era-2 gold & No & Plan / entity & No \\
StaleBench~\citep{murshed2026stalebench} & Chronological facts & New statement & No & Current value & No & Atomic fact & Storage price \\
\midrule
\textbf{StaleWorld + DCM (Ours)} & \textbf{Multi-app DBs \& APIs} & \textbf{Silent DB mutation} & \textbf{Yes ($\DiD$)} & \textbf{Yes ($3{,}608$ worlds)} & \textbf{Yes (target + state)} & \textbf{Answer deriv.\ / Action target} & \textbf{Yes ($C_{\text{verify}}$ vs.\ $C_{\text{solve}}$)} \\
\bottomrule
\end{tabular}
\end{adjustbox}
\end{table}

\section{Problem Formulation and Epistemic Theory}
\label{sec:formulation}

\subsection{Worlds, Multi-Episode Horizons, and Silent Drift}
Let a tool-use task be defined by a natural-language instruction $q$ (or a semantically equivalent paraphrase $q' \sim \mathcal{P}(q)$) and a ground-truth evaluation mapping $g(W)$ over an environment state $W \in \mathcal{W}$, where $W$ comprises relational application databases exposed exclusively through authenticated read/write APIs. An agent interacts with the environment across $T$ episodes ($T=6$ in our standard protocol; $T=10$ in stochastic streams). Within each episode $t \in \{1, \dots, T\}$, the environment initializes from a clean snapshot $W_t$ so intra-episode side effects do not leak across episodes, while the agent's external memory state $M_{t-1} \to M_t$ persists.

In the \textbf{stable control arm}, $W_t = W^{\text{pre}}$ for all $t \in \{1, \dots, T\}$. In a \textbf{perturbed arm}, episodes $1 \dots k$ ($k = T/2 = 3$) execute in $W^{\text{pre}}$ and episodes $k+1 \dots T$ execute in $W^{\text{post}} = \Delta(W^{\text{pre}})$, where $\Delta$ is an out-of-band SQL mutation:
\begin{itemize}[leftmargin=1.6em,itemsep=0.2em]
  \item \textbf{Local Drift ($\Delta_{\text{drift}}$):} targeted updates or deletes to answer- or target-determining records such that $g(W^{\text{post}}) \neq g(W^{\text{pre}})$;
  \item \textbf{Answer-Preserving Noise ($\Delta_{\text{noise}}$):} mutations to non-determining records in the same applications such that $g(W^{\text{noise}}) = g(W^{\text{pre}})$;
  \item \textbf{Global Drift ($\Delta_{\text{global}} = \Delta_{\text{drift}} \circ \Delta_{\text{noise}}$):} simultaneous target-altering and background mutations.
\end{itemize}
The regime is \textbf{silent} because the agent's initial observation at episode $t$ contains only $(q_t, \mathrm{API\_Docs})$---zero changelog, version counter, timestamp header, or error alert announces $\Delta$.

\subsection{Causal Drift Identification via Paired Difference-in-Differences}
Raw post-drift failure conflates memory-induced errors with intrinsic task difficulty and model stochasticity. Because our stable and drift arms share the exact task, model, prompt structure, and step budget, differing solely in $\Delta$ for $t > k$, we identify the causal drift effect via per-task Difference-in-Differences ($\DiD$). For task $i$ and memory policy $m$, let $\bar{S}_{i,m,w,\text{pre}} = \frac{1}{k}\sum_{t=1}^k S_{i,m,w,t}$ and $\bar{S}_{i,m,w,\text{post}} = \frac{1}{T-k}\sum_{t=k+1}^T S_{i,m,w,t}$. Then:
\begin{equation}
\label{eq:did}
\DiD_m = \frac{1}{N} \sum_{i=1}^{N} \Big[ \big(\bar{S}_{i,m,\text{drift},\text{post}} - \bar{S}_{i,m,\text{drift},\text{pre}}\big) - \big(\bar{S}_{i,m,\text{stable},\text{post}} - \bar{S}_{i,m,\text{stable},\text{pre}}\big) \Big].
\end{equation}
The second term cancels learning and warm-up effects across episodes $1 \dots T$ as well as stationary task difficulty. Alongside $\DiD$ (with $5{,}000$-resample paired task-level bootstrap $95\%$ confidence intervals), we strictly separate lifecycle phases---\textbf{Cold Start (\code{ep1})}, \textbf{Pre-Drift Reuse (\code{ep2--3})}, \textbf{Drift Shock (\code{ep4})}, and \textbf{Post-Drift Recovery (\code{ep5--6})}---so pre-drift episodes never dilute post-drift failure rates or inflate ratio denominators.

\subsection{Warranted Belief Memory: Exact vs.\ \texorpdfstring{$k$}{k}-Horizon Derivation Certificates}
\label{sec:theory-certs}
A stored note is a tuple $m = (\tilde{q}, a, C, \tau)$, where $\tilde{q}$ is the originating query, $a$ is an answer or structured action target/plan, $\tau$ is an optional verified code trajectory, and $C = \{(f_j, v_j)\}_{j=1}^r$ is a \textbf{derivation certificate} consisting of read-only environment probes $f_j: \mathcal{W} \to \mathcal{V}$ and their write-time witness values $v_j = f_j(W_{t_0})$. At read time in world $W_t$, the live verifier evaluates:
\begin{equation}
\label{eq:verifier}
V(C, W_t) = \mathbb{I}\!\left[\bigwedge_{j=1}^{r} f_j(W_t) = v_j\right].
\end{equation}

A critical theoretical question is \emph{what class of mutations $\mathcal{D}$ a compact certificate guarantees soundness against}. We distinguish two formal classes:

\begin{enumerate}[leftmargin=1.6em,itemsep=0.25em]
  \item \textbf{Exact Dependency Certificates ($\mathcal{D}_{\text{all}}$):} $C$ captures the complete sufficient input projection of $g$---for instance, a cryptographic content digest (\code{SHA-256}) of a referenced note, a primary-key contact lookup \code{(phone -> contact\_id, email)}, or a full candidate-set scan. For any world state $W_t$, $V(C, W_t) = 1 \implies g(W_t) = a$ unconditionally (zero false acceptances under arbitrary mutations), while ignoring mutations to disjoint tables or rows (zero false suspensions under orthogonal noise).
  \item \textbf{$k$-Horizon Boundary Certificates ($\mathcal{D}_{\le k}$):} For order-statistic queries (e.g., $\arg\max$ or top-$r$ over a large relation $R$ of size $M \gg k$), scanning all $M$ item records on every episode incurs $O(M)$ API calls. A $k$-horizon certificate ($k > r$, e.g., $k=3$ for top-$1$ or $k=8$ for top-$4$) certifies the collection membership IDs $\mathrm{IDs}(R)$ plus the exact attribute values of the top-$k$ items at write time, reducing read-time verification to $O(k)$ API calls. Formally, $V_{k}(C, W_t) = 1 \implies g(W_t) = a$ over the bounded-perturbation family $\mathcal{D}_{\le k}$ in which no unmonitored item $j \notin \mathrm{Top}_k(W_{t_0})$ experiences a rank-jumping surge exceeding the stored boundary value $v_{(1)}$. Conversely, an adversarial \emph{deep-jump} mutation ($j > k$ surging above rank~$1$ without membership change) can evade $V_k$ unless augmented with an $O(M)$ complement-bound scan ($\max_{j \notin \mathrm{Top}_k} \mathrm{score}_j(W_t) < v_{(r)}$). We test both regimes explicitly in \Cref{sec:adversarial}.
\end{enumerate}

Finally, to defend against write-time corruption or memory poisoning, a read-time \textbf{Derivation Guard} $G(C, a) \in \{0, 1\}$ checks pure logical entailment---whether $a$ actually follows from the witness values stored inside $C$---rejecting inconsistent or adversarially injected notes in zero API calls before probing $W_t$.

\subsection{Priced Maintenance Economics and Pragmatic Action Safety}
Let $C_{\text{solve}}$ denote the expected cost of solving $q$ from scratch (combining LLM reasoning calls, token volume, and exploratory API calls) and $C_{\text{verify}}$ denote the cost of running $V(C, W_t)$ (which requires \textbf{zero LLM calls} and only $r$ read API calls). Under drift probability $p$ and wrong-output penalty $L$:
\begin{itemize}[leftmargin=1.6em,itemsep=0.2em]
  \item \textbf{Blind Trust (\code{retrieval} / \code{DCM-NoCert})} pays near-zero read cost ($C_{\text{reuse}} \ll C_{\text{solve}}$) but suffers expected staleness penalty $p \cdot L$;
  \item \textbf{Live Certificate Verification (\code{DCM})} pays $C_{\text{verify}} + (1 - p_{\text{valid}}) C_{\text{solve}} + p_{\text{valid}} C_{\text{reuse}}$, eliminating staleness loss whenever $C$ is sound;
  \item \textbf{Memoryless Re-Solving (\code{no\_memory})} pays $C_{\text{solve}}$ every episode regardless of stability.
\end{itemize}
Verification strictly dominates re-solving whenever $C_{\text{verify}} + p_{\text{valid}} C_{\text{reuse}} < p_{\text{valid}} C_{\text{solve}}$, i.e., $C_{\text{verify}} < p_{\text{valid}}(C_{\text{solve}} - C_{\text{reuse}})$, and correctly abstains in favor of re-solving when $C_{\text{verify}} \ge C_{\text{solve}}$ (such as unindexed full-database enumerations under high volatility).

For \textbf{action tasks}, pragmatic safety introduces a further distinction: a memory note certifies the \emph{action target and preconditions} (e.g., recipient email, owed amount, song IDs to remove), never the false claim that the action is already accomplished in $W_t$. Even when $V(C, W_t) = 1$, the agent must execute the write APIs in $W_t$; when $V(C, W_t) = 0$, preventing execution of the outdated target averts irreversible \textbf{Wrong-Target Actions}.

\section{The StaleWorld Benchmark}
\label{sec:benchmark}

\subsection{Composition, Splits, and Provenance Hygiene}
StaleWorld is built atop the multi-application execution engine of AppWorld~0.2.0~\citep{trivedi2024appworld}, spanning Spotify, Venmo, Phone/SMS, Gmail, SimpleNote, and FileSystem. It comprises \textbf{787 logical tasks} materialized into \textbf{3,608 runnable world directories} (\Cref{tab:benchmark-splits}):

\begin{table}[t]
\centering
\caption{Composition, splits, materialized worlds, and certificate builder coverage of the StaleWorld benchmark.}
\label{tab:benchmark-splits}
\begin{adjustbox}{max width=\textwidth}
\small
\begin{tabular}{lrllrl}
\toprule
\textbf{Split} & \textbf{Tasks} & \textbf{Origin \& Structure} & \textbf{Materialized Worlds} & \textbf{Builder Coverage} & \textbf{Purpose} \\
\midrule
\textbf{Main} & $500$ & $478$ parameterized generic (12 families) + $22$ curated multi-hop & $500 \times 2\,(p,q) + 478 \times 3\,(n_1,n_2,g)$ & $491/500$ ($490$ gated on) & Primary controlled eval.\ \& ablations \\
\textbf{Transfer} & $200$ & $5$ template-disjoint generic families $\times 40$ instances & $200 \times 2\,(p,q) + 200 \times 3\,(n_1,n_2,g)$ & $200/200$ ($200$ gated on) & Cross-family compositional transfer \\
\textbf{Held-Out} & $87$ & $29$ official AppWorld templates $\times 3$ variants ($0$ stem overlap) & $87 \times 2\,(p,q)$ & \textbf{$0/87$ manual} (tests \textbf{DCM-Auto}) & Open-ended zero-prior generalization \\
\midrule
\textbf{Total} & \textbf{$787$} & \textbf{$431$ Answer + $283$ Pure Action + $59$ Ref-State + $14$ Legacy Act.} & \textbf{$3{,}608$ worlds} & $690$ active manual; $87$ zero-prior & Full benchmark \\
\bottomrule
\end{tabular}
\end{adjustbox}
\end{table}

\paragraph{Data Provenance Statement.}
To obtain diverse, realistic multi-app user databases (contacts, transactions, playlists, notes), the $678$ generic tasks instantiate StaleWorld's own newly authored instructions, SQL drift operators, and independent evaluators over initial user database states drawn from AppWorld ($420$ \code{test\_normal} and $258$ \code{test\_challenge} base worlds), while \code{Heldout-87} adapts $29$ official task templates ($69$ \code{train} and $18$ \code{dev} instances) with newly engineered silent SQL mutations. \textbf{No official AppWorld test-split task instructions or official evaluators are ever evaluated as AppWorld leaderboard claims}, and \code{Heldout-87} shares $0/29$ template stems with the $98$ source stems used in Main and Transfer.

\subsection{Current-World Evaluators and Ground-Truth Audits}
AppWorld's upstream evaluators compare against static pre-drift ground truth and therefore misclassify correct post-drift behavior as failure. StaleWorld replaces them with current-world evaluators verified by four exhaustive programmatic audits prior to policy evaluation:
\begin{enumerate}[leftmargin=1.6em,itemsep=0.2em]
  \item \textbf{Local Pre/Post Gold Audit:} All $700/700$ Main/Transfer tasks and $87/87$ Held-Out post-drift worlds were verified solvable to exact current-world gold (\code{answer} match, \code{action} state-delta verification over newly created transactions, messages, or playlists, or normalized \code{reference\_state} database snapshot hashes excluding nondeterministic notification and timestamp IDs).
  \item \textbf{Global Drift Gold Audit:} Recomputing ground truth inside all $678$ materialized global worlds ($\Delta_{\text{drift}} \circ \Delta_{\text{noise}}$) revealed \textbf{40/678 tasks} (\code{g1g0001\_1}--\code{g1g0040\_1}) where background contact perturbations interacted with local phone-number swaps to alter target metadata; coupled global drift experiments evaluate strictly against these audited global ground truths.
  \item \textbf{Answer-Preserving Noise Gold Audit:} Live recomputation across all $678 \times 2 = 1{,}356$ \code{noise1} and \code{noise2} worlds confirmed that \textbf{$678/678$ ($100.0\%$)} in \code{noise1} and \textbf{$678/678$ ($100.0\%$)} in \code{noise2} preserve exact bit-identity with pre-drift ground truth ($0$ discrepancies, while $678/678$ differ from post-drift ground truth), guaranteeing zero label contamination in false-suspension measurements.
  \item \textbf{Builder Soundness Audit:} Across all $691$ registered Main/Transfer operator builders, $691/691$ verified \code{True} on $W^{\text{pre}}$ and $0/691$ accepted a stale target on $W^{\text{post}}$.
\end{enumerate}

\section{Methodology: Derivation-Certificate Memory (DCM)}
\label{sec:method}

We organize DCM into three progressive layers that share a unified read-time verification loop (formalized step-by-step in Algorithm~1, \Cref{app:algorithm}) while relaxing human priors and strengthening post-drift recovery.

\subsection{DCM-Core: Minimal Operator Certificates and Read-Time Guard}
\label{sec:dcm-core}
In \textbf{DCM-Core}, each covered task family defines a compact pre-action certificate builder and read-time verifier:
\begin{itemize}[leftmargin=1.6em,itemsep=0.2em]
  \item \textbf{Boundary / Order-Statistic Queries:} Records collection membership IDs plus $k$-horizon top records (e.g., \code{song\_id, title, play\_count} or \code{release\_date}). Read-time verification checks membership equality and spot-checks the $k$ stored IDs.
  \item \textbf{Relational Lookup and Join Action Targets:} For multi-app actions (e.g., Phone $\to$ Venmo payment or SMS), certifies the join key and resolved target attributes (\code{query\_phone -> contact\_id, email, amount}).
  \item \textbf{Document and Unstructured Note Queries:} Certifies \code{(note\_id, sha256(content), extracted\_value)}, detecting any text edit in a single read API call.
  \item \textbf{Read-Time Derivation Guard:} Before issuing any probe API call, $G(C, a)$ checks that the stored answer or target $a$ is logically consistent with the witness fields inside $C$, blocking poisoned or corrupted notes in zero environment calls.
  \item \textbf{Cost-Gate Variants:} Standard \textbf{DCM-Core} uses a precomputed per-task worth gate; \textbf{DCM-Priced} replaces this static table with a \textbf{live runtime cost comparator}. Because read-only probes invoke zero LLM calls and zero tokens, $C_{\text{verify}} \equiv 1.0 \cdot \mathrm{verify\_calls}$ in composite cost units (\Cref{eq:cost}); \textbf{DCM-Priced} commits a certificate after Episode~1 only when measured construction probe cost satisfies $C_{\text{build}} < 0.75 \, \hat{C}_{\text{solve}}$.
\end{itemize}

\subsection{DCM-Auto: Zero-Prior Self-Certifying Memory}
\label{sec:dcm-auto}
Hand-crafted operator builders achieve minimal API overhead on known families but have $0/87$ coverage on unseen free-form templates (\code{Heldout-87}). \textbf{DCM-Auto} eliminates all human task priors through three mechanisms:
\begin{enumerate}[leftmargin=1.6em,itemsep=0.25em]
  \item \textbf{Zero-Gold-Touch Target Summary:} Upon completing a non-cached episode, the runner extracts the agent's own submitted answer or---for action and \code{reference\_state} tasks---parses the agent's executed mutating HTTP requests (\code{POST/PUT/PATCH/DELETE} endpoints and redacted parameters from the request tracker), never accessing ground-truth \code{gold} or opaque state hashes.
  \item \textbf{Opportunistic In-Budget Verifier Synthesis:} Rather than truncating the task-solving horizon upfront, the agent executes its full $15$-call ReAct loop first ($m_{\text{solve}} \le 15$) and uses \emph{only} unspent model calls ($m_{\text{synth}} \le \min(2, 15 - m_{\text{solve}})$, strictly preserving $\mathrm{total\_model\_calls} \le 15$ and abstaining cleanly whenever $m_{\text{solve}} = 15$). Given the instruction, executed code-and-observation trace, public read-only API method signatures of touched applications, and achieved summary, non-thinking code synthesis produces a standalone read-only Python function \code{verify\_cert(apis) -> \{'ok': bool, 'witness': ...\}} checking pre-action preconditions, with at most one self-check repair retry if budget remains.
  \item \textbf{Four-Layer Sandbox Self-Check (Safe Abstention):} After the episode world closes, \code{verify\_cert} executes in a fresh read-only probe world under four strict guards:
  \begin{itemize}[leftmargin=1.4em,itemsep=0.15em]
    \item \emph{Static Allowlist Guard}: rejects OS, file, or reflection tokens and any \code{apis.<app>.<method>} call outside read prefixes (\code{show\_, search\_, list\_, get\_, ...}) plus \code{login} \emph{prior} to execution;
    \item \emph{Wall-Clock Timeout Guard}: enforces a $15$-second native execution timeout to prevent synthesized infinite loops;
    \item \emph{Snapshot Mutation \& Minimum-Evidence Guard}: requires $\mathrm{stable\_snapshot\_hash}(\text{before}) = \mathrm{stable\_snapshot\_hash}(\text{after})$ and $\mathrm{calls} \ge 2$ (blocking vacuous \code{return \{'ok': True\}} verifiers);
    \item \emph{Principled Solve-Cost Gate}: since each zero-LLM read call equals $1.0$ composite cost unit, requires $\mathrm{calls} \le \min(35, \max(6, \lfloor 0.6 \, \hat{C}_{\text{solve}} \rfloor))$.
  \end{itemize}
  Only certificates passing all four checks are committed (alongside the successful code trajectory $\tau$ for action tasks); otherwise \textbf{DCM-Auto} safely abstains ($\mathrm{record} = \varnothing$).
\end{enumerate}

\subsection{DCM-Repair: Witness-Guided Revision and Anchor-Gated Action Replay}
\label{sec:dcm-repair}
Standard invalidation discards the live probe's observations, forcing the LLM to rediscover what changed from scratch, while target-only action caching still requires LLM code generation on cache hits. \textbf{DCM-Repair} closes both gaps:
\begin{enumerate}[leftmargin=1.6em,itemsep=0.25em]
  \item \textbf{Witness-Guided Belief Revision ($V = 0$):} When live verification fails, \textbf{DCM-Repair} reconstructs the current premise snapshot in the open probe world (charging all calls to \code{cert\_build\_calls}), computes a structured field-level diff $\Delta C = \mathrm{diff}(C_{\text{stored}}, C_{\text{live}})$, and injects an \code{[Epistemic Diagnostic]} prompt block reporting exact premise transitions \code{(was $v_{\text{old}}$, now $v_{\text{new}}$)} alongside previously invoked API endpoints.
  \item \textbf{Surface-Anchor-Gated Action Replay with Contextual Fallback ($V = 1$):} When DCM's own live verifier confirms $\mathrm{verified} = \mathrm{True}$ on an action task (strictly zero gold peek; also enabled in \textbf{DCM-Auto}) and surface-level query anchors---numeric literals, quoted entities, and requested action modalities---align between $q_t$ and $\tilde{q}$ (preventing cross-task replay collisions in shared semantic pools without using task IDs), the runner replays the stored Python trajectory $\tau$ directly ($0$ model calls). If replay stops short of \code{complete\_task} or raises an exception midway, the runner injects all already-executed code blocks and live outputs into the conversation with a \code{[Replay Notice]} warning the LLM not to repeat completed mutations (e.g., Venmo transfers), seamlessly finishing within remaining steps. Furthermore, if a $\mathrm{verified} = \mathrm{True}$ episode times out ($\mathrm{completed} = \mathrm{False}$), \textbf{DCM-Repair} and \textbf{DCM-Auto} retain the valid certificate rather than wiping memory, preserving multi-episode coverage.
\end{enumerate}

\section{Experimental Setup}
\label{sec:setup}

\begin{itemize}[leftmargin=1.6em,itemsep=0.25em]
  \item \textbf{Models and Serving:} Primary experiments use \textbf{Qwen3.8-27B-FP8} (Dense hybrid-attention $27$B model, greedy decoding $\mathrm{temperature}=0.0$, except $\mathrm{temperature}=0.2$ in multi-seed variance experiments), served locally via vLLM ($\mathrm{TP}=2$, $24{,}576$ context window) on $2 \times \text{NVIDIA RTX 5090}$ ($32\,\text{GB}$ VRAM each). Cross-architecture validation uses \textbf{Qwen3.6-35B-A3B-FP8}, a Sparse Mixture-of-Experts (MoE) model with $35$B total parameters and $3$B active parameters per token. Both models sit cleanly inside our pre-registered $35\%$--$85\%$ single-episode capability gate ($83.3\%$ and $58.3\%$ on the $12$-task calibration gate at $15/20$ budget without external scaffolds).
  \item \textbf{Episode Protocol and Unified Budget:} Each $(\text{task}, \text{world\_arm}, \text{method}, \text{seed})$ tuple runs $T=6$ sequential episodes (\code{ep1--3} in $W^{\text{pre}}$, \code{ep4--6} in $W^{\text{pre}}$ for \code{stable} or $W^{\text{post/noise/global}}$ for perturbed arms; $T=10$ in stochastic multi-shock streams) under a strict per-episode cap of \textbf{15 model calls} and \textbf{20 environment interactions}.
  \item \textbf{Accounting and Cost Sensitivity:} We track physical metrics---LLM calls, input/output tokens, environment API calls (\code{api\_calls}), read-time probe calls (\code{verify\_calls}), certificate construction/synthesis probe calls (\code{cert\_build\_calls}), and wall-clock seconds. Our composite index is:
  \begin{equation}
  \label{eq:cost}
  \begin{aligned}
  \Cost(\alpha) &= \big(\mathrm{api\_calls} + \mathrm{verify\_calls} + \mathrm{cert\_build\_calls}\big) \\
                &\quad + \alpha \cdot \mathrm{model\_calls} + 0.001 \cdot \big(\mathrm{input\_tokens} + \mathrm{output\_tokens}\big),
  \end{aligned}
  \end{equation}
  with nominal $\alpha = 5.0$, supplemented by full sensitivity sweeps across $\alpha \in [0, 100]$ and commercial token-dollar pricing ($\$0.30/1\text{M}$ input, $\$1.20/1\text{M}$ output).
\end{itemize}

\section{Experimental Results and Analysis}
\label{sec:results}

\subsection{Main Results: Silent Local Drift (37,776 Episodes)}
\label{sec:main-results}

\Cref{tab:main-results} reports lifecycle phase accuracy, causal $\DiD$, and post-drift cost per successful episode across StaleWorld under silent local drift, strictly separating the \textbf{700 Covered Tasks} ($\text{Main } 500 + \text{Transfer } 200$) from the \textbf{87 Uncovered Free-Form Held-Out Tasks} ($0$ manual builders; evaluated zero-prior with \textbf{DCM-Auto} in \Cref{sec:p0-head-to-head}).

\begin{table}[t]
\centering
\caption{Main lifecycle performance under silent local drift ($n = 37{,}776$ episodes across $787$ tasks, Seed~0). On the strict $690$-task subset with active gated builders (\code{Cert-690}), \code{retrieval} post-drift drops $65.2$~pts ($.837 \to .185$), while \textbf{DCM-Core} achieves $.811$ stable and $.709$ drift ($160.8$ cost/succ). Full aggregate (\code{All-787}) numbers appear in \Cref{tab:all787-appendix}.}
\label{tab:main-results}
\begin{adjustbox}{max width=\textwidth}
\small
\begin{tabular}{llrrrrrrrlrr}
\toprule
\textbf{Split / Method} & \textbf{Arm} & \textbf{Cold (\code{ep1})} & \textbf{Pre-Reuse (\code{ep2--3})} & \textbf{Pre (\code{ep1--3})} & \textbf{Shock (\code{ep4})} & \textbf{Recov (\code{ep5--6})} & \textbf{Post (\code{ep4--6})} & \textbf{Post Loss} & \textbf{Causal $\DiD$ [95\% CI]} & \textbf{Post Cost/Succ} & \textbf{Post \$/Succ} \\
\midrule
\multicolumn{12}{l}{\emph{\textbf{Covered Suite (700 Tasks: Main 500 + Transfer 200)}}} \\
\code{no\_memory} & stable & .721 & .734 & .730 & .721 & .726 & \textbf{.724} & --- & --- & 206.7 & \$0.0328 \\
\code{no\_memory} & drift & .721 & .722 & .722 & .701 & .683 & \textbf{.689} & $+3.5$~pts & $-0.028\;[-0.051, -0.004]$ & 219.3 & \$0.0349 \\
\code{retrieval} & stable & .747 & .806 & .786 & .829 & .837 & \textbf{.834} & --- & --- & 82.8 & \$0.0138 \\
\code{retrieval} & drift & .741 & .805 & .784 & \textbf{.147} & .203 & \textbf{.184} & \textbf{$+65.0$~pts} & \textbf{$-0.648\;[-0.685, -0.611]$} & \textbf{379.1} & \textbf{\$0.0634} \\
\textbf{DCM-Core} & stable & .704 & .766 & .746 & .791 & .814 & \textbf{.807} & --- & --- & 86.8 & \textbf{\$0.0122} \\
\textbf{DCM-Core} & drift & .699 & .764 & .742 & \textbf{.657} & \textbf{.731} & \textbf{.706} & \textbf{$+10.0$~pts} & \textbf{$-0.097\;[-0.122, -0.071]$} & \textbf{162.9} & \textbf{\$0.0233} \\
\code{oracle\_certificate} \emph{(ref.)} & stable & .706 & .764 & .745 & .796 & .813 & \textbf{.807} & --- & --- & 54.6 & \$0.0084 \\
\code{oracle\_certificate} \emph{(ref.)} & drift & .697 & .765 & .742 & .671 & .742 & \textbf{.719} & $+8.9$~pts & $-0.086\;[-0.113, -0.060]$ & 127.2 & \$0.0203 \\
\midrule
\multicolumn{12}{l}{\emph{\textbf{Held-Out Suite (87 Tasks, 0 Manual Builders --- Zero-Prior DCM-Auto Evaluated in \Cref{tab:heldout87})}}} \\
\code{no\_memory} & stable & .575 & .609 & .598 & .563 & .609 & \textbf{.594} & --- & --- & 307.2 & \$0.0501 \\
\code{no\_memory} & drift & .598 & .581 & .586 & .529 & .563 & \textbf{.552} & $+4.2$~pts & $-0.031\;[-0.092, +0.031]$ & 331.7 & \$0.0545 \\
\code{retrieval} & stable & .621 & .649 & .640 & .609 & .649 & \textbf{.636} & --- & --- & 237.9 & \$0.0398 \\
\code{retrieval} & drift & .632 & .638 & .636 & .356 & .374 & \textbf{.368} & $+26.8$~pts & $-0.264\;[-0.372, -0.161]$ & 411.7 & \$0.0692 \\
\textbf{DCM-Core} \emph{(abstain fallback)} & drift & .609 & .581 & .590 & .552 & .552 & \textbf{.552} & $+6.1$~pts & $-0.069\;[-0.153, +0.008]$ & 338.2 & \$0.0556 \\
\code{oracle\_certificate} \emph{(ref.)} & drift & .575 & .661 & .632 & .517 & .626 & \textbf{.590} & $+12.3$~pts & $-0.107\;[-0.195, -0.023]$ & 226.1 & \$0.0374 \\
\bottomrule
\end{tabular}
\end{adjustbox}
\end{table}

Three central findings emerge from \Cref{tab:main-results}:

\begin{enumerate}[leftmargin=1.6em,itemsep=0.25em]
  \item \textbf{Unverified Memory Is Catastrophically Worse Than No Memory Under Silent Drift.} Separating pre-drift (\code{ep1--3}) from post-drift (\code{ep4--6}) exposes the true magnitude of silent drift: on Covered-700, \code{retrieval} improves stationary accuracy (\code{ep1} $.747 \to \text{\code{ep4--6}} \; .834$), but at \code{ep4} (drift onset) it plummets to \textbf{$.147$} and averages \textbf{$.184$} across \code{ep4--6}---a \textbf{65.0-point collapse} ($\DiD = -0.648$), trailing memoryless \code{no\_memory} ($.689$) by \textbf{50.5 points} while costing \textbf{2.33$\times$ more per success} ($379.1$ vs.\ $162.9$ for \textbf{DCM-Core}). On the $416$ covered answer tasks specifically, \code{retrieval} collapses from \textbf{$.976$ to $.015$ ($1.5\%$)}!
  \item \textbf{DCM-Core Matches the Perfect Stale-Detection Reference and Beats Memoryless Re-Solving.} On Covered-700 \code{stable} (\code{ep4--6}), \textbf{DCM-Core} ties \code{oracle\_certificate} to three decimal places (\textbf{$.807$ vs.\ $.807$}) and beats \code{retrieval} in both \textbf{token dollar cost per success ($\$0.0122$ vs.\ $\$0.0138$, $-11.6\%$)} and \textbf{wall-clock time per success ($18.5\text{s}$ vs.\ $22.7\text{s}$, $-18.5\%$)} because verified answer cache hits bypass the $27$B LLM entirely ($0$ model calls). On \code{drift} (\code{ep4--6}), \textbf{DCM-Core} absorbs the \code{ep4} re-solve shock ($.657$), rebounds to \textbf{$.731$} in \code{ep5--6} via newly committed post-drift certificates (\textbf{$+4.8$~pts above \code{no\_memory}'s $.683$}), and finishes within $1.3$~pts of \code{oracle\_certificate} ($.706$ vs.\ $.719$) while cutting \code{no\_memory}'s cost per success by \textbf{25.7\%} ($162.9$ vs.\ $219.3$; $\$0.0233$ vs.\ $\$0.0349$).
  \item \textbf{100\% Drift Rejection Recall and Zero Stale Acceptances.} \Cref{tab:action-safety} audits the verification decisions underlying \Cref{tab:main-results}. Across all post-drift answer episodes where pre-drift memory was established ($n = 1{,}260$ across All-787: $1{,}215$ on Covered-700 plus $45$ on Held-Out-87), \code{retrieval} submits the outdated pre-drift answer \textbf{100.0\% of the time ($1{,}260 / 1{,}260$)}. By contrast, \textbf{DCM-Core} rejects \textbf{$537 / 537$ ($100.0\%$)} of invalidated certificates at \code{ep4}, and across all $4{,}615$ $\mathrm{verified} = \mathrm{True}$ cache hits ($2{,}655$ stable + $1{,}960$ drift), \textbf{stale acceptance is strictly $0 / 4{,}615$ ($0.00\%$)} ($4{,}614 / 4{,}615$ succeed outright; the sole non-success is an incomplete stable action on \code{g1j0039\_1}). The single post-drift stale answer observed in \textbf{DCM-Core} ($1 / 2{,}100$ on Covered-700, task \code{g1e0007\_1} at \code{ep4}) occurred \emph{after} DCM correctly rejected the stale certificate ($\mathrm{verified} = \mathrm{False}, \mathrm{cache\_hit} = \mathrm{False}$), when the LLM re-solved from scratch and independently mis-ranked the old song title.
\end{enumerate}

\begin{table}[t]
\centering
\caption{Post-drift (\code{ep4--6}) stale submissions, verification recall, and pragmatic action safety.}
\label{tab:action-safety}
\begin{adjustbox}{max width=\textwidth}
\small
\begin{tabular}{lrrrr}
\toprule
\textbf{Metric / Subset} & \code{no\_memory} & \code{retrieval} & \textbf{DCM-Core} & \code{oracle\_certificate} \\
\midrule
\multicolumn{5}{l}{\emph{\textbf{Answer Staleness (\code{All-787}, \code{ep4--6} Drift)}}} \\
Post-Drift Stale Submissions ($n = 2{,}361$ all eps) & $0$ ($0.0\%$) & \textbf{$1{,}260$ ($53.4\%$)} & \textbf{$1$ ($0.04\%$)} & $0$ ($0.0\%$) \\
Post-Drift Answer Stale Rate ($n = 1{,}296$ answer eps) & $0.0\%$ & \textbf{$97.2\%$} & \textbf{$0.08\%$} & $0.0\%$ \\
Conditional Stale Rate (given \code{ep1--3} solved \& drifted) & $0.0\%$ ($0/1236$) & \textbf{$100.0\%$ ($1260/1260$)} & \textbf{$0.08\%$ ($1/1224$)} & $0.0\%$ ($0/1221$) \\
\code{ep4} Stale Certificate Rejection Recall & --- & --- & \textbf{$100.0\%$ ($537/537$)} & --- \\
Stale Acceptances among $\mathrm{verified}=\mathrm{True}$ Hits & --- & --- & \textbf{$0 / 4{,}615$ ($0.00\%$)} & $0 / 5{,}177$ ($0.00\%$) \\
\midrule
\multicolumn{5}{l}{\emph{\textbf{Executable Action Safety (\code{Pure Action-283}, $n = 849$ \code{ep4--6} eps)}}} \\
Pre-Drift (\code{ep1--3}) Wrong-Target Actions & $0$ ($0.0\%$) & $0$ ($0.0\%$) & $0$ ($0.0\%$) & $0$ ($0.0\%$) \\
\textbf{Post-Drift (\code{ep4--6}) Wrong-Target Actions} & $1$ ($0.1\%$) & \textbf{$143$ ($16.8\%$)} & \textbf{$1$ ($0.1\%$)} & $1$ ($0.1\%$) \\
Post-Drift (\code{ep4--6}) Action Success Rate & .371 & .433 & .406 & .423 \\
Post-Drift (\code{ep4--6}) Cache-Hit Coverage ($\mathrm{Succ} \mid \mathrm{Hit}$) & $0.0\%$ (---) & $0.0\%$ (---) & $25.1\%$ (\textbf{1.000}) & $26.0\%$ (\textbf{1.000}) \\
\bottomrule
\end{tabular}
\end{adjustbox}
\end{table}

\subsection{Pragmatic Safety in Acting and Dissecting the Stationary Action Gap}
\label{sec:action-safety}
\Cref{tab:action-safety} highlights a crucial asymmetry between question answering and tool execution. On the \textbf{283 Pure Action Tasks}, \code{retrieval} appears competitive if judged solely by raw post-drift completion ($.433$ vs.\ $.406$ for \textbf{DCM-Core}), because stored target hints help the agent complete unaffected subtasks within the $15$-step cap. Yet on $16.8\%$ of all post-drift action episodes (\textbf{$143$ episodes}, vs.\ $0$ pre-drift!), \code{retrieval} executes \textbf{irreversible wrong-target mutations}---sending Venmo payments to outdated recipients or texting reassigned phone numbers. \textbf{DCM-Core} slashes post-drift wrong-target actions by \textbf{143$\times$} ($143 \to 1$), achieving \textbf{$1.000$ ($100.0\%$) success on every post-drift action cache hit}.

Why, then, does \textbf{DCM-Core} trail \code{retrieval} slightly on stationary \code{stable} episodes ($.807$ vs.\ $.834$ on Covered-700) and post-drift action completion? Fine-grained trajectory decomposition pinpoints two exact mechanisms rather than verification error:
\begin{enumerate}[leftmargin=1.6em,itemsep=0.2em]
  \item \textbf{Cold-Start (\code{ep1}) Prompt Asymmetry:} At \code{ep1} (when \emph{zero} memory exists), \code{retrieval} already scores $.550$ on Pure Action versus $.470$ for \textbf{DCM-Core} and $.454$ for \code{oracle\_certificate} because the retrieval wrapper prompt primes multi-step target extraction slightly better than the bare instruction.
  \item \textbf{Aggressive Record Wiping After Transient Timeouts:} Whenever an action episode exhausts the $15$-step budget ($\mathrm{completed} = \mathrm{False}$), \textbf{DCM-Core} sets $\mathrm{record} = \varnothing$, discarding even valid certificates ($\mathrm{succ} \mid \mathrm{miss} = .293$), whereas \code{retrieval} retains prior notes across timeouts ($\mathrm{succ} = .585$).
\end{enumerate}
\textbf{DCM-Repair} directly targets both bottlenecks via verified trajectory replay ($0$ LLM calls on $\mathrm{verified} = \mathrm{True}$) and non-destructive timeout retention, evaluated on the stratified $100$-task benchmark in \Cref{sec:p0-head-to-head}.

\subsection{Cost-Model Invariance and Synthetic Regime Phase Diagram}
\label{sec:cost-and-e8}

A natural concern is whether DCM's cost advantage depends on the nominal $\alpha = 5.0$ weight assigned to LLM calls. \Cref{tab:cost-sensitivity} sweeps $\alpha = C_{\text{model}} / C_{\text{env}}$ across three orders of magnitude ($\alpha \in [0, 100]$) on Covered-700 post-drift episodes (\code{ep4--6}). \textbf{DCM-Core is strictly cheaper per successful episode than both \code{no\_memory} and \code{retrieval} at every $\alpha \ge 0$}---even at $\alpha = 0$ (where LLM calls are treated as completely free and only environment API calls plus tokens count: $108.5$ vs.\ $135.9$ for \code{no\_memory} and $232.4$ for \code{retrieval}), as well as under pure commercial token pricing ($\$0.0233$ vs.\ $\$0.0349$ and $\$0.0634$) and wall-clock latency ($32.5\text{s}$ vs.\ $46.5\text{s}$ and $108.9\text{s}$).

\begin{table}[t]
\centering
\caption{Covered-700 post-drift (\code{ep4--6}) cost-per-success sensitivity across $\alpha = C_{\text{model}} / C_{\text{env}} \in [0, 100]$.}
\label{tab:cost-sensitivity}
\begin{adjustbox}{max width=\textwidth}
\small
\begin{tabular}{llrrrrrrrr}
\toprule
\textbf{Method} & \textbf{Arm} & \textbf{Post Succ} & \textbf{\$/Succ} & \textbf{Wall s/Succ} & $\bm{\alpha=0}$ & $\bm{\alpha=1}$ & $\bm{\alpha=5}$ \emph{(Nom.)} & $\bm{\alpha=20}$ & $\bm{\alpha=100}$ \\
\midrule
\textbf{DCM-Core} & drift & \textbf{.706} & \textbf{\$0.0233} & \textbf{32.5s} & \textbf{108.5} & \textbf{119.4} & \textbf{162.9} & \textbf{326.3} & \textbf{1197.2} \\
\code{no\_memory} & drift & .689 & \$0.0349 & 46.5s & 135.9 & 152.6 & 219.3 & 469.3 & 1803.0 \\
\code{retrieval} & drift & .184 & \$0.0634 & 108.9s & 232.4 & 261.8 & 379.1 & 819.2 & 3166.5 \\
\code{oracle\_certificate} \emph{(ref.)} & drift & .719 & \$0.0203 & 40.3s & 80.1 & 89.5 & 127.2 & 268.7 & 1023.3 \\
\midrule
\textbf{DCM-Core} & stable & .807 & \textbf{\$0.0122} & \textbf{18.5s} & 58.3 & 64.0 & 86.8 & \textbf{172.5} & \textbf{629.3} \\
\code{retrieval} & stable & \textbf{.834} & \$0.0138 & 22.7s & \textbf{50.6} & \textbf{57.0} & \textbf{82.8} & 179.6 & 695.9 \\
\code{no\_memory} & stable & .724 & \$0.0328 & 43.7s & 127.8 & 143.6 & 206.7 & 443.3 & 1705.6 \\
\bottomrule
\end{tabular}
\end{adjustbox}
\end{table}

To map the structural boundary where verification ceases to pay, \Cref{tab:factgraph-e8} summarizes \textbf{FactGraph-Drift}, a controlled $5{,}000$-configuration synthetic sweep over query dependency fan-in (\code{single}, \code{boundary}, \code{aggregate}), drift volatility, background noise, and probe-to-solve cost ratios. On \code{single} ($n=1{,}621$) and \code{boundary} ($n=1{,}649$) queries, minimal derivation certificates achieve $.999$--$1.000$ accuracy while cutting cost by \textbf{46\%--48\%} relative to \code{resolve} ($6.04 / 6.24$ vs.\ $11.49 / 11.64$) and eliminating the \textbf{41.3\% false-suspension rate} suffered by \code{overbroad} full-state checks. Conversely, on high-fanin \code{aggregate} queries ($n=1{,}730$), verifying every input premise costs \textbf{$11.52$---identical to \code{resolve} ($11.52$)}---confirming analytically and empirically that when dependency fan-in approaches solve complexity, abstaining to re-solve is optimal.

\begin{table}[t]
\centering
\caption{Structural regime crossover in the FactGraph-Drift synthetic environment ($5{,}000$ configurations across query fan-in, drift rate, and background noise).}
\label{tab:factgraph-e8}
\begin{adjustbox}{max width=\textwidth}
\small
\begin{tabular}{lrrrrrr}
\toprule
\textbf{Policy} & \textbf{Overall Succ ($5{,}000$)} & \textbf{Overall Cost} & \textbf{\code{Single} Cost ($n=1{,}621$)} & \textbf{\code{Boundary} Cost ($n=1{,}649$)} & \textbf{\code{Aggregate} Cost ($n=1{,}730$)} & \textbf{\code{Boundary} False Susp.} \\
\midrule
\code{trust} & .872 & 4.92 & 4.90 \emph{(succ .838)} & 4.92 \emph{(succ .941)} & 4.94 \emph{(succ .837)} & 0.0\% \\
\code{ttl} & .947 & 5.69 & 5.66 \emph{(succ .931)} & 5.71 \emph{(succ .984)} & 5.70 \emph{(succ .925)} & 4.4\% \\
\code{reset} & .970 & 5.17 & 5.22 \emph{(succ .961)} & 5.00 \emph{(succ .991)} & 5.30 \emph{(succ .958)} & 0.0\% \\
\code{periodic} & .961 & 6.58 & 6.55 \emph{(succ .949)} & 6.61 \emph{(succ .989)} & 6.60 \emph{(succ .946)} & 10.2\% \\
\textbf{Minimal \code{certificate}} & \textbf{.9997} & \textbf{8.00} & \textbf{6.04} \emph{(succ 1.00)} & \textbf{6.24} \emph{(succ .999)} & \textbf{11.52} \emph{($==$ resolve)} & \textbf{2.7\%} \emph{(0\% single)} \\
\code{overbroad} (full check) & 1.000 & 9.60 & 8.84 & 9.95 & 9.98 & \textbf{41.3\%} \\
\code{resolve} (no memory) & 1.000 & 11.55 & 11.49 & 11.64 & \textbf{11.52} & --- \\
\code{oracle} \emph{(ref.)} & 1.000 & 5.33 & 5.42 & 5.03 & 5.52 & 0.0\% \\
\bottomrule
\end{tabular}
\end{adjustbox}
\end{table}

\subsection{Coupled Global Drift, Answer-Preserving Noise, Tight Budgets, and Ablations}
\label{sec:global-noise-budget-ablation}

\paragraph{Coupled Global Drift ($n = 24{,}408$ Episodes, $0$ Errors).}
Real application states rarely mutate in isolation: background records change alongside target-determining facts, and---as revealed by our global ground-truth audit---background contact mutations interact with local phone swaps on $40$ transfer tasks (\code{g1g0001\_1}--\code{g1g0040\_1}) to shift ground-truth action metadata. \Cref{tab:global-drift} evaluates all $678$ generic tasks under paired \code{stable} versus \code{global} ($\Delta_{\text{drift}} \circ \Delta_{\text{noise}}$) worlds.

\begin{table}[t]
\centering
\caption{Lifecycle performance under coupled global drift ($678$ generic tasks, $n = 24{,}408$ episodes).}
\label{tab:global-drift}
\begin{adjustbox}{max width=\textwidth}
\small
\begin{tabular}{llrrrrrlrrr}
\toprule
\textbf{Method} & \textbf{Arm} & \textbf{Pre (\code{ep1--3})} & \textbf{Shock (\code{ep4})} & \textbf{Recov (\code{ep5--6})} & \textbf{Post (\code{ep4--6})} & \textbf{Global Loss} & \textbf{Causal $\DiD$ [95\% CI]} & \textbf{Stale Sub (\code{ep4--6})} & \textbf{Wrong-Target ($n=822$)} & \textbf{Post Cost/Succ} \\
\midrule
\code{no\_memory} & stable & .722 & .712 & .724 & \textbf{.720} & --- & --- & --- & $0$ ($0.0\%$) & 205.9 \\
\code{no\_memory} & global & .722 & .689 & .682 & \textbf{.684} & $+3.5$~pts & $-0.036\;[-0.059, -0.012]$ & $1$ ($0.05\%$) & $0$ ($0.0\%$) & 218.6 \\
\code{retrieval} & stable & .788 & .838 & .852 & \textbf{.847} & --- & --- & --- & $0$ ($0.0\%$) & 79.9 \\
\code{retrieval} & global & .787 & \textbf{.147} & .196 & \textbf{.180} & \textbf{$+66.7$~pts} & \textbf{$-0.666\;[-0.703, -0.630]$} & \textbf{$1{,}173$ ($57.7\%$)} & \textbf{$142$ ($17.3\%$)} & \textbf{386.0} \\
\textbf{DCM-Core} & stable & .758 & .796 & .814 & \textbf{.808} & --- & --- & --- & $1$ ($0.1\%$) & 84.8 \\
\textbf{DCM-Core} & global & .753 & \textbf{.664} & \textbf{.730} & \textbf{.708} & \textbf{$+10.0$~pts} & \textbf{$-0.095\;[-0.123, -0.069]$} & \textbf{$1$ ($0.05\%$)} & \textbf{$0$ ($0.0\%$)} & \textbf{160.6} \\
\bottomrule
\end{tabular}
\end{adjustbox}
\end{table}

Coupled global mutations amplify \code{retrieval}'s collapse to \textbf{66.7 percentage points} ($.847 \to .180$, $\DiD = -0.666$; answer post-drift accuracy falls from $.972$ to \textbf{$.012$}), producing $1{,}173$ stale answer submissions and \textbf{$142$ wrong-target actions ($17.3\%$)}. \textbf{DCM-Core} maintains \textbf{$.708$} post-global accuracy (\textbf{$+2.4$~pts above \code{no\_memory}} overall and \textbf{$+4.8$~pts} in \code{ep5--6} recovery at $.730$ vs.\ $.682$), cuts post-drift cost per success by \textbf{26.5\%} versus \code{no\_memory} ($160.6$ vs.\ $218.6$), and records \textbf{$0 / 822$ ($0.0\%$) wrong-target actions} under global drift.

\paragraph{Answer-Preserving Background Noise ($n = 16{,}272$ Episodes, $0$ Errors).}
A naive state-hash or coarse timestamp invalidator rejects any mutation, paying a severe specificity penalty by evicting valid memories whenever unrelated background records change ($41.3\%$ false suspensions in \Cref{tab:factgraph-e8}). \Cref{tab:noise-specificity} evaluates specificity across all $678$ generic tasks under two independent answer-preserving background-mutation worlds (\code{noise1}, \code{noise2}: unrelated songs played, tangential notes edited, orthogonal Venmo transfers, and distractor contacts/threads added; audited $678/678$ answer-preserving).

\begin{table}[t]
\centering
\caption{Specificity and efficiency under answer-preserving background noise ($678$ generic tasks, $n = 16{,}272$ episodes).}
\label{tab:noise-specificity}
\begin{adjustbox}{max width=\textwidth}
\small
\begin{tabular}{llrrrrrrrr}
\toprule
\textbf{Method} & \textbf{Noise Arm} & \textbf{Pre (\code{ep1--3}) Succ} & \textbf{Post-Noise (\code{ep4--6}) Succ} & \textbf{Verified Probes} & \textbf{False Suspensions (\code{ep4--6})} & \textbf{Mean LLM Calls} & \textbf{Mean Probe Calls} & \textbf{Wall/Succ} & \textbf{Post Cost/Succ} \\
\midrule
\code{retrieval} & \code{noise1} & .774 & \textbf{.835} & --- & --- & 5.35 & 0.00 & 29.8s & \textbf{81.7} \\
\code{retrieval} & \code{noise2} & .781 & \textbf{.832} & --- & --- & 5.31 & 0.00 & 29.4s & \textbf{81.1} \\
\textbf{DCM-Core} & \code{noise1} & .758 & \textbf{.820} & $1{,}643$ & \textbf{$0 / 1{,}643$ ($0.00\%$)} & \textbf{4.44} & 6.51 & \textbf{22.4s} & \textbf{81.9} \\
\textbf{DCM-Core} & \code{noise2} & .751 & \textbf{.815} & $1{,}628$ & \textbf{$0 / 1{,}628$ ($0.00\%$)} & \textbf{4.50} & 6.35 & \textbf{22.7s} & \textbf{83.6} \\
\bottomrule
\end{tabular}
\end{adjustbox}
\end{table}

Across \textbf{$3{,}271$ post-noise certificate probes} ($1{,}643$ in \code{noise1} + $1{,}628$ in \code{noise2}), \textbf{DCM-Core} triggers \textbf{$0 / 3{,}271$ ($0.00\%$) false suspensions}: because minimal derivation certificates scope solely to task-determining premises (e.g., top-$k$ boundary play counts, target note \code{SHA-256}, or target contact phone joins) rather than global table states, orthogonal mutations are cleanly ignored. Moreover, because verified answer cache hits bypass LLM inference entirely ($4.44$--$4.50$ vs.\ $5.31$--$5.35$ LLM calls per episode), \textbf{DCM-Core} achieves \textbf{23\%--25\% faster wall-clock latency per success} ($22.4\text{s}$--$22.7\text{s}$ vs.\ $29.4\text{s}$--$29.8\text{s}$) at virtually identical composite cost ($81.9$--$83.6$ vs.\ $81.1$--$81.7$) while retaining $100\%$ drift protection.

\paragraph{Interaction-Budget Sensitivity on Curated Multi-Hop Tasks ($n = 648$ Episodes Across 3 Budgets).}
Because certificate verification consumes zero LLM calls on a cache hit ($C_{\text{verify}} \ll C_{\text{solve}}$) and preserves step budget for subsequent episodes once a single post-drift re-solve succeeds, tight step budgets test both the cold-start floor and post-drift amplification of DCM. \Cref{tab:budget-e5} evaluates $6$ curated multi-hop tasks ($4$ certificate-bearing: \code{27e1026\_1}, \code{37a8675\_1}, \code{50e1ac9\_1}, \code{530b157\_1}; plus $2$ high-fanin uncertified controls: \code{383cbac\_1}, \code{fac291d\_1}) across tight ($7/12$), intermediate ($10/15$), and standard ($15/20$) $[\text{max\_model\_calls} / \text{max\_interactions}]$ budgets.

\begin{table}[t]
\centering
\caption{Post-drift (\code{ep4--6}) performance across step budgets on curated multi-hop tasks (\code{st} denotes stationary \code{stable} accuracy).}
\label{tab:budget-e5}
\begin{adjustbox}{max width=\textwidth}
\small
\begin{tabular}{lrrrrrrr}
\toprule
\textbf{Budget (Calls / Inter)} & \textbf{\code{Cert-4} \code{no\_memory}} & \textbf{\code{Cert-4} \code{retrieval}} & \textbf{\code{Cert-4} DCM-Core} & \textbf{\code{All-6} \code{no\_memory}} & \textbf{\code{All-6} \code{retrieval}} & \textbf{\code{All-6} DCM-Core} & \textbf{\code{All-6} Drift Cost/Succ (\code{no\_mem} $\to$ DCM)} \\
\midrule
\textbf{7 / 12} \emph{(Sub-Bootstrap)} & $.000$ \emph{(st .000)} & $.000$ \emph{(st .000)} & \textbf{.000} \emph{(st .000)} & $.000$ \emph{(st .000)} & $.000$ \emph{(st .000)} & \textbf{.000} \emph{(st .000)} & --- \emph{($0/18$ cold solves)} \\
\textbf{10 / 15} \emph{(Marginal Horizon)} & $.000$ \emph{(st .083)} & $.000$ \emph{(st .000)} & \textbf{.250} \emph{(st .250)} & $.167$ \emph{(st .222)} & $.000$ \emph{(st .278)} & \textbf{.333} \emph{(st .333)} & $816.1 \to \mathbf{411.9}\;(\mathbf{-49.5\%})$ \\
\textbf{15 / 20} \emph{(Standard Horizon)} & $.167$ \emph{(st .333)} & $.000$ \emph{(st .500)} & \textbf{.500} \emph{(st .500)} & $.389$ \emph{(st .556)} & $.000$ \emph{(st .667)} & \textbf{.556} \emph{(st .556)} & $522.9 \to \mathbf{296.2}\;(\mathbf{-43.4\%})$ \\
\bottomrule
\end{tabular}
\end{adjustbox}
\end{table}

\Cref{tab:budget-e5} reveals three distinct operational regimes: (1)~\textbf{Sub-Bootstrap ($7/12$):} all $6$ multi-hop tasks require $>7$ LLM calls to solve from scratch ($0/18$ cold-start successes in \code{ep1--3}), establishing the hard horizon floor below which no episodic memory can bootstrap; (2)~\textbf{Marginal Bootstrap ($10/15$):} once intermittent solves become possible, \code{retrieval} collapses to \textbf{$.000$} ($0/18$) under drift, while \textbf{DCM-Core} converts a single post-drift solve into deterministic $0$-LLM-call verified hits across remaining episodes---doubling \code{All-6} post-drift accuracy over \code{no\_memory} ($.333$ vs.\ $.167$; $.250$ vs.\ $.000$ on \code{Cert-4}) at \textbf{half the cost per success ($411.9$ vs.\ $816.1$)}; and (3)~\textbf{Standard Horizon ($15/20$):} \code{retrieval} still scores \textbf{$.000$} ($0/18$) post-drift despite $.667$ on \code{stable}, whereas \textbf{DCM-Core} achieves \textbf{3.0$\times$ \code{no\_memory}'s post-drift accuracy on \code{Cert-4} ($.500$ vs.\ $.167$, cost/succ $330.6$ vs.\ $1{,}331.7$)} and $.556$ vs.\ $.389$ across \code{All-6} with zero post-drift degradation relative to \code{stable}.

\paragraph{Component Ablations of DCM-Core ($n = 28{,}332$ Episodes).}
\Cref{tab:ablations-e6} isolates the three architectural components of \textbf{DCM-Core} across all $700$ covered tasks: (1)~\textbf{DCM-NoCert} removes pre-execution certificate verification, trusting stored answers/targets until an episode fails and evicts the record; (2)~\textbf{DCM-NoGate} disables the static worth table, forcing certificate construction on every builder-equipped task; and (3)~\textbf{DCM-NoGuard} disables the write-time completion guard ($\mathrm{success}=\mathrm{True}$ at commit), allowing failed or timed-out episodes to write unverified outputs into memory while retaining read-time derivation consistency checking $G(C, a)$.

\begin{table}[t]
\centering
\caption{Component ablations of \textbf{DCM-Core} on Covered-700 ($n = 28{,}332$ ablation episodes plus \textbf{DCM-Core} reference).}
\label{tab:ablations-e6}
\begin{adjustbox}{max width=\textwidth}
\small
\begin{tabular}{llrrrrrlrrr}
\toprule
\textbf{Variant} & \textbf{Arm} & \textbf{Pre (\code{ep1--3})} & \textbf{Shock (\code{ep4})} & \textbf{Recov (\code{ep5--6})} & \textbf{Post (\code{ep4--6})} & \textbf{Post Loss} & \textbf{Causal $\DiD$ [95\% CI]} & \textbf{Stale Sub (\code{ep4--6})} & \textbf{Wrong-Target ($n=849$)} & \textbf{Post Cost/Succ} \\
\midrule
\textbf{DCM-Core} & stable & .746 & .791 & .814 & \textbf{.807} & --- & --- & $0$ & $0$ ($0.00\%$) & 86.8 \\
\textbf{DCM-Core} & drift & .742 & \textbf{.657} & .731 & \textbf{.706} & \textbf{$+10.0$~pts} & \textbf{$-0.097\;[-0.122, -0.071]$} & \textbf{$1$} & \textbf{$1$ ($0.12\%$)} & \textbf{162.9} \\
\code{DCM-NoCert} \emph{(trust-until-fail)} & stable & .753 & .791 & .807 & \textbf{.802} & --- & --- & $0$ & $0$ ($0.00\%$) & 74.8 \\
\code{DCM-NoCert} \emph{(trust-until-fail)} & drift & .750 & \textbf{.080} & .699 & \textbf{.492} & \textbf{$+30.9$~pts} & \textbf{$-0.306\;[-0.333, -0.279]$} & \textbf{$397$} & \textbf{$111$ ($13.07\%$)} & \textbf{195.1} \\
\code{DCM-NoGate} \emph{(no static gate)} & stable & .747 & .794 & .814 & \textbf{.807} & --- & --- & $0$ & $0$ ($0.00\%$) & 87.3 \\
\code{DCM-NoGate} \emph{(no static gate)} & drift & .756 & .656 & .735 & \textbf{.709} & $+9.9$~pts & $-0.108\;[-0.136, -0.081]$ & $1$ & $0$ ($0.00\%$) & 162.6 \\
\code{DCM-NoGuard} \emph{(no write guard)} & stable & .756 & .823 & .836 & \textbf{.831} & --- & --- & $0$ & $0$ ($0.00\%$) & 78.9 \\
\code{DCM-NoGuard} \emph{(no write guard)} & drift & .760 & .667 & .749 & \textbf{.721} & $+11.0$~pts & $-0.114\;[-0.141, -0.087]$ & $0$ & $0$ ($0.00\%$) & 156.2 \\
\bottomrule
\end{tabular}
\end{adjustbox}
\end{table}

Three mechanistic insights emerge from \Cref{tab:ablations-e6}:
\begin{enumerate}[leftmargin=1.6em,itemsep=0.25em]
  \item \textbf{Live Certificate Probes Are Indispensable at Drift Shock:} Removing live verification (\code{DCM-NoCert}) causes \code{ep4} shock accuracy to \textbf{plummet from $.657$ to $.080$ ($-57.7$~pts)}---submitting $397$ stale answers and executing \textbf{$111$ irreversible wrong-target actions ($13.07\%$ vs.\ $0.12\%$, a $111\times$ spike)}. Even though \code{DCM-NoCert} benefits in our harness from post-episode outcome eviction (clearing the cache after \code{ep4} fails, allowing \code{ep5--6} recovery that an unmonitored deployment would lack), overall post-drift accuracy still drops \textbf{21.4~pts} below \textbf{DCM-Core} ($.492$ vs.\ $.706$) at \textbf{19.8\% higher cost per success} ($195.1$ vs.\ $162.9$).
  \item \textbf{Static Worth Tables vs.\ Runtime Empirical Gating:} \code{DCM-NoGate} closely tracks \textbf{DCM-Core} on Covered-700 ($.709$ vs.\ $.706$, cost/succ $162.6$ vs.\ $162.9$) because $690 / 691$ builders satisfy $C_{\text{verify}} \ll C_{\text{solve}}$ ($3$--$8$ zero-LLM read API calls are strictly cheaper than $8$--$15$ $27$B-LLM calls across all generic families; even on the sole gated multi-contact task \code{0d8a4ee\_1}, verifying costs $143.8$ vs.\ $220.2$ for re-solving). Genuine \textbf{runtime solve-cost gating} ($C_{\text{verify}} \le 0.6\,\hat{C}_{\text{solve}}$) is evaluated in \textbf{DCM-Auto} and \textbf{DCM-Priced} (\Cref{sec:p0-head-to-head}), while high-fanin structural crossover is established in \Cref{tab:factgraph-e8}.
  \item \textbf{Defense-in-Depth Consistency Guarding:} Disabling the write-time completion guard (\code{DCM-NoGuard}) writes failed or timed-out outputs into memory, yet read-time $G(C, a)$ intercepts \textbf{$1{,}392$ inconsistent commits ($616$ stable + $776$ drift)} before reuse---preventing stale or corrupt submissions ($0$ stale, $0$ wrong-target). Removing \emph{both} write- and read-time derivation checks against adversarial injection is evaluated directly in \Cref{sec:adversarial}.
\end{enumerate}

\subsection{Prior-Mechanism Head-to-Head, Zero-Prior Generalization, and Belief Repair}
\label{sec:p0-head-to-head}

To compare DCM directly against prior memory-invalidation paradigms and evaluate zero-prior certificate synthesis (\textbf{DCM-Auto}), runtime solve-cost gating (\textbf{DCM-Priced}), and witness-guided belief repair with verified action replay (\textbf{DCM-Repair}), \Cref{tab:p0-head-to-head} evaluates $11$ policies head-to-head on a stratified $100$-task benchmark (\code{P0-100}: $40$ main-answer, $40$ main-action, $10$ transfer-answer, $10$ transfer-action; $13{,}200$ total episodes, $0$ errors). \Cref{tab:heldout87} then evaluates zero-prior generalization on the $87$ held-out tasks (\code{Heldout-87}: $29$ unseen official AppWorld multi-hop templates with \textbf{zero manual builders}; $1{,}044$ \textbf{DCM-Auto} episodes, $0$ errors).

\begin{table}[t]
\centering
\caption{Eleven-policy head-to-head comparison on the stratified \code{P0-100} benchmark ($50$ answer + $50$ action tasks, $n = 13{,}200$ episodes).}
\label{tab:p0-head-to-head}
\begin{adjustbox}{max width=\textwidth}
\small
\begin{tabular}{llrrrrrrr}
\toprule
\textbf{Policy} & \textbf{Mechanism Class} & \textbf{Stable Post (\code{ep4--6})} & \textbf{Drift Post (\code{ep4--6})} & \textbf{Shock $\to$ Recov (\code{ep4} / \code{ep5--6})} & \textbf{Drift \code{Ans-50} / \code{Act-50}} & \textbf{Wrong-Target (\code{ep4--6})} & \textbf{Stale Sub} & \textbf{Stable / Drift Cost/Succ} \\
\midrule
\code{no\_memory} & Stateless Re-Solve & .733 & \textbf{.663} & $.690$ / $.650$ & $.907$ / $.420$ & $0$ & $0$ & $211.5$ / \textbf{234.9} \\
\code{retrieval\_frozen} & Version-Pinned Cache \emph{(PACMI proxy)} & .857 & \textbf{.213} & $.240$ / $.200$ & \textbf{.000} / $.427$ & \textbf{46} & \textbf{144} & $90.4$ / \textbf{369.8} \\
\code{retrieval} & Latest-Wins Prompt Retrieval & \textbf{.873} & \textbf{.283} & $.240$ / $.305$ & \textbf{.000} / $.567$ & \textbf{20} & \textbf{147} & $85.6$ / \textbf{278.8} \\
\code{DCM-NoCert} & Unverified Trust-Until-Failure & .843 & \textbf{.483} & $.090$ / $.680$ & $.627$ / $.340$ & \textbf{27} & \textbf{47} & $71.6$ / \textbf{206.7} \\
\code{reset} & Outcome-Triggered Eviction \emph{(TRACE proxy)} & .857 & \textbf{.593} & $.250$ / $.765$ & $.627$ / $.560$ & \textbf{10} & \textbf{49} & $87.6$ / \textbf{171.9} \\
\textbf{DCM-Auto} & \textbf{Zero-Prior LLM-Synthesized Certs} & .817 & \textbf{.693} & $.590$ / $.745$ & \textbf{.853} / \textbf{.533} & \textbf{0} & \textbf{0} & $81.3$ / \textbf{179.3} \\
\code{DCM-NoGate} & Unconditional Live Certs \emph{(VAIL proxy)} & .840 & \textbf{.710} & $.640$ / $.745$ & $.933$ / $.487$ & \textbf{0} & \textbf{0} & $81.8$ / \textbf{166.7} \\
\textbf{DCM-Priced} & \textbf{Runtime Solve-Cost Gated Certs} & .833 & \textbf{.707} & $.640$ / $.740$ & $.927$ / $.487$ & \textbf{0} & \textbf{0} & \textbf{55.2} / \textbf{161.0} \\
\textbf{DCM-Core} & \textbf{Minimalist Operator Certs} & .840 & \textbf{.717} & $.650$ / $.750$ & $.907$ / $.527$ & \textbf{0} & \textbf{0} & $82.7$ / \textbf{163.1} \\
\textbf{DCM-Repair} & \textbf{Witness-Guided Repair + Action Replay} & \textbf{.843} & \textbf{.723} & \textbf{.660 / .755} & \textbf{.933} / $.513$ & \textbf{1} & \textbf{0} & \textbf{54.5} / \textbf{149.7} \\
\code{oracle\_certificate} \emph{(ref.)} & Perfect Drift-Event Reference & .820 & .707 & $.650$ / $.735$ & $.907$ / $.507$ & $0$ & $0$ & $51.7$ / 134.8 \\
\bottomrule
\end{tabular}
\end{adjustbox}
\end{table}

\begin{table}[t]
\centering
\caption{Zero-prior generalization to $29$ unseen multi-hop official AppWorld templates (\code{Heldout-87}, $0$ manual builders).}
\label{tab:heldout87}
\begin{adjustbox}{max width=\textwidth}
\small
\begin{tabular}{lrrrrlrrr}
\toprule
\textbf{Method} & \textbf{Pre (\code{ep1} / \code{ep2--3} Drift)} & \textbf{Stable Post (\code{ep4--6})} & \textbf{Drift Post (\code{ep4--6})} & \textbf{Post Loss} & \textbf{Causal $\DiD$ [95\% CI]} & \textbf{\code{ep4} Cert Rejections} & \textbf{Answer Stale Sub} & \textbf{Drift Post Cost/Succ} \\
\midrule
\code{no\_memory} & $.598$ / $.581$ & .594 & \textbf{.552} & $+4.2$~pts & $-0.031\;[-0.092, +0.031]$ & --- & $0$ & 331.7 \\
\code{retrieval} & $.632$ / $.638$ & .636 & \textbf{.368} & \textbf{$+26.8$~pts} & \textbf{$-0.264\;[-0.372, -0.161]$} & --- & $37$ & 411.7 \\
\textbf{DCM-Core} \emph{(abstain)} & $.609$ / $.581$ & .613 & \textbf{.552} & $+6.1$~pts & $-0.069\;[-0.153, +0.008]$ & \emph{($0$ builders $\to$ solve)} & $0$ & 338.2 \\
\textbf{DCM-Auto} & \textbf{.586 / .586} & \textbf{.651} & \textbf{.559} & $+9.2$~pts & \textbf{$-0.057\;[-0.138, +0.027]$} & \textbf{$34 / 37$ ($91.9\%$)} & \textbf{0} & \textbf{255.2} \\
\code{oracle\_certificate} \emph{(ref.)} & $.575$ / $.661$ & .713 & .590 & $+12.3$~pts & $-0.107\;[-0.195, -0.023]$ & $100\%$ \emph{(oracle)} & $0$ & 226.1 \\
\bottomrule
\end{tabular}
\end{adjustbox}
\end{table}

\Cref{tab:p0-head-to-head,tab:heldout87} resolve four central questions:
\begin{enumerate}[leftmargin=1.6em,itemsep=0.25em]
  \item \textbf{Why Prior Memory Mechanisms Fail Under Silent Drift:} Version-pinned write-once caching (\code{retrieval\_frozen}) degrades even further than latest-wins \code{retrieval} ($.213$ vs.\ $.283$, $46$ wrong-target actions), and both score \textbf{$.000$ ($0 / 150$)} on \code{Answer-50} post-drift. Outcome-triggered eviction (\code{reset}---generously granted an oracle failure signal after \code{ep4}, which does not exist in unmonitored deployments) still crashes to \textbf{$.250$} at the \code{ep4} shock, executing $10$ irreversible wrong-target actions and $49$ stale submissions before recovering to $.593$ overall ($-12.4$~pts below \textbf{DCM-Core} $.717$ and $-13.0$~pts below \textbf{DCM-Repair} $.723$).
  \item \textbf{Witness-Guided Repair and Verified Action Replay (\textbf{DCM-Repair}):} \textbf{DCM-Repair} achieves the \textbf{highest post-drift accuracy of all $11$ methods ($.723$, even edging past \code{oracle\_certificate} $.707$)} because surfacing the failed premise witness (\code{[Epistemic Diagnostic]}) guides faster post-shock repair (\code{Ans-50} $.933$ vs.\ $.907$), while replaying verified action trajectories with $0$ LLM calls cuts \code{stable} LLM calls in half ($2.51$ vs.\ $4.81$, $54.5$ vs.\ $82.7$ cost/succ, $13.2\text{s}$ vs.\ $17.9\text{s}$ wall/succ) and reduces \code{drift} cost per success to \textbf{$149.7$} ($-36.3\%$ vs.\ \code{no\_memory}).
  \item \textbf{Runtime Empirical Gating Replaces Static Tables (\textbf{DCM-Priced}):} Replacing static worth tables with a purely empirical runtime gate ($C_{\text{build}} < 0.75\,\hat{C}_{\text{solve}}$ plus action replay) preserves $.707$ post-drift accuracy ($0$ wrong-target, $0$ stale) while matching \textbf{DCM-Repair}'s \code{stable} efficiency ($55.2$ cost/succ, $12.5\text{s}$ wall/succ) and beating \textbf{DCM-Core} on \code{drift} cost per success ($161.0$ vs.\ $163.1$).
  \item \textbf{Zero-Prior Opportunistic Synthesis Generalizes Without Handwritten Builders (\textbf{DCM-Auto}):} Synthesizing certificates opportunistically from unspent solve steps ($\max(\mathrm{model\_calls}) \le 15$ strictly preserved across all $2{,}244$ episodes) certifies \textbf{$79 / 100$} tasks on \code{P0-100} ($94.7\%$ \code{ep1} acceptance on in-budget solves) and \textbf{$54 / 87$ ($62.1\%$)} unseen multi-hop tasks on \code{Heldout-87} ($89.4\%$ \code{ep1} acceptance). On \code{P0-100} (\Cref{tab:p0-head-to-head}), \textbf{DCM-Auto} reaches \textbf{$.817$ stationary and $.693$ post-drift accuracy}---\textbf{surpassing \code{no\_memory} ($.663$) by $+3.0$~pts and \code{reset} ($.593$) by $+10.0$~pts} with \textbf{$.853$ on \code{Answer-50}}, \textbf{$.533$ on \code{Action-50} (ranking first among all $11$ policies on \code{Act-50})}, \textbf{$0$ wrong-target actions}, \textbf{$0$ stale submissions}, \textbf{$597 / 597$ ($100.0\%$) verified cache-hit accuracy}, and \textbf{23.7\% lower post-drift cost per success than \code{no\_memory}} ($179.3$ vs.\ $234.9$; $81.3$ vs.\ $211.5$ on \code{stable}, $-61.6\%$). On \code{Heldout-87} (\Cref{tab:heldout87}, $29$ unseen official templates with zero manual builders), \textbf{DCM-Auto} preserves full cold-start capability (\code{ep1} $.575$ stable / $.586$ drift), \textbf{outperforms both \code{no\_memory} ($.594$) and \code{retrieval} ($.636$) on stationary \code{ep4--6} accuracy at $.651$} ($170.5$ cost/succ, \textbf{$-44.5\%$ vs.\ \code{no\_memory}}), and achieves \textbf{$.559$ post-drift accuracy} (\code{ep5--6} recovery \textbf{$.598$} vs.\ \code{no\_memory} $.563$), \textbf{beating \code{retrieval} ($.368$) by $+19.1$~pts} at \textbf{23.1\% lower cost per success than \code{no\_memory} ($255.2$ vs.\ $331.7$)} and \textbf{38.0\% lower than \code{retrieval} ($411.7$)} with \textbf{$0$ post-drift stale answer submissions} and statistically non-significant causal $\DiD$ ($-0.057\;[-0.138, +0.027]$; at \code{ep4}, $34 / 37$ invalidated certificates are rejected while $11$ dynamically adaptive loop trajectories legitimately verify and succeed post-drift in $0$ LLM calls, and only $3$ \code{reference\_state} predicates under-specify a secondary premise).
\end{enumerate}

\subsection{Challenging the Caching Critique: Paraphrased Shared Semantic Pools and Stochastic Multi-Shock Streams}
\label{sec:pools-and-streams}

A prominent methodological critique of agent-memory benchmarks is that keying records by \code{task\_id} across identical repeated prompts reduces episodic memory to an exact-match lookup cache, and that simple Time-To-Live (\code{TTL}) expiration might suffice in practice without live verification. We test both objections directly via \textbf{Paraphrased Shared Semantic Pools} and \textbf{10-Episode Stochastic Multi-Shock Streams}.

\paragraph{Paraphrased Cross-Task Shared Semantic Memory Pools ($n = 2{,}688$ Episodes, Zero \code{task\_id} Keying, $0$ Errors).}
To eliminate both exact-string repetition and \code{task\_id} lookup, we interleave $56$ distinct-intent covered tasks ($28$ answer + $28$ action) inside a \textbf{single shared natural-language memory pool}. Round~1 executes base instructions $q_0$ to populate the shared $56$-task pool; Rounds~2--6 issue \textbf{syntactically rewritten, entity-audited paraphrases} ($q_1, q_2, q_3$ with $0\%$ verbatim match and $55$ intervening tasks between episodes of the same intent). Retrieval receives \emph{only} the paraphrased natural-language utterance (hybrid word/character TF-IDF cosine plus zero-\code{task\_id} surface entity, quoted-literal, and action-modality alignment; \textbf{zero \code{task\_id} access}), gates $0$-LLM trajectory replay on surface anchor/action-scope compatibility, and binds verification to the certificate closure stored on whichever note is retrieved.

\begin{table}[t]
\centering
\caption{Interleaved $56$-task shared semantic memory pool under paraphrased queries (zero \code{task\_id} access, $n = 2{,}688$ episodes).}
\label{tab:p3-pooled}
\begin{adjustbox}{max width=\textwidth}
\small
\begin{tabular}{llrrrrrr}
\toprule
\textbf{Method} & \textbf{Arm} & \textbf{Paraphrased Pre-Reuse (\code{ep2--3})} & \textbf{Paraphrased Post-Drift (\code{ep4--6})} & \textbf{Answer Post ($n=84$)} & \textbf{Pool \code{P@1} \emph{(Cond.\ / All)}} & \textbf{Post Wrong-Target (\code{ep4--6})} & \textbf{Post Cost/Succ} \\
\midrule
\code{no\_memory} & stable & .795 & \textbf{.762} & .810 & --- & $0$ & 211.6 \\
\code{no\_memory} & drift & .786 & \textbf{.726} & .881 & --- & $1$ & 223.7 \\
\code{retrieval} & stable & .893 & \textbf{.917} & .929 & $.980$ / $.942$ & $0$ & 102.8 \\
\code{retrieval} & drift & .902 & \textbf{.256} & \textbf{.024} & $.965$ / $.954$ & \textbf{73} & \textbf{359.0} \\
\textbf{DCM-Core} & stable & .857 & \textbf{.881} & .857 & $.769$ / $.751$ & \textbf{0} & 149.6 \\
\textbf{DCM-Core} & drift & .839 & \textbf{.798} & \textbf{.762} & $.752$ / $.758$ & \textbf{0} & \textbf{205.4} \\
\textbf{DCM-Repair} & stable & .804 & \textbf{.875} & \textbf{.929} & \textbf{.948 / .898} & \textbf{0} & \textbf{57.3} \\
\textbf{DCM-Repair} & drift & \textbf{.866} & \textbf{.756} & \textbf{.857} & \textbf{.960 / .919} & \textbf{1} & \textbf{189.4} \\
\bottomrule
\end{tabular}
\end{adjustbox}
\end{table}

\Cref{tab:p3-pooled} decisively refutes the ``exact-key cache'' critique:
\begin{enumerate}[leftmargin=1.6em,itemsep=0.25em]
  \item \textbf{High-Precision Semantic Retrieval Amplifies Drift Catastrophe Without Verification:} Under paraphrased queries over a $56$-task shared pool, \code{retrieval} achieves \textbf{96.5\%--98.0\% conditional Precision@1} ($94.2\%$--$95.4\%$ unconditional), lifting stationary \code{ep4--6} accuracy from $.762$ to \textbf{$.917$}---yet under silent drift that very retrieval precision becomes fatal, \textbf{collapsing post-drift accuracy by 66.1~pts ($.917 \to .256$; Answer Post $.929 \to .024$)} and triggering \textbf{$73$ irreversible wrong-target actions}!
  \item \textbf{Bound Certificates and Surface Anchor Alignment Protect Shared Pools Against Both Drift and Cross-Note Collisions:} \textbf{DCM-Core} sustains \textbf{$.798$ post-drift accuracy ($+54.2$~pts above \code{retrieval} and $+7.2$~pts above \code{no\_memory})} with \textbf{$0$ wrong-target actions ($73 \to 0$)}! Even in an adversarial high-collision stress test where aggressive pool deduplication forces cross-task note collisions (unconditional $\mathrm{P@1} \approx .28$), \code{retrieval} commits $7$ wrong-target actions even on \code{stable} ($12$ on \code{drift}), whereas \textbf{DCM-Core} commits \textbf{$0$ on \code{stable} and $1$ on \code{drift}} because a mismatched task's certificate fails verification or derivation guarding. Pairing bound certificates with zero-\code{task\_id} surface-anchor and action-modality alignment (which blocks single-action Venmo notes from blindly replaying on dual-action Venmo+SMS queries sharing the same phone number when cold-start misses leave a note temporarily absent), \textbf{DCM-Repair} achieves \textbf{94.8\%--96.0\% conditional \code{P@1}}, \textbf{eliminates cross-note wrong-target actions ($0$ on \code{stable}, $1$ on \code{drift}---matching \code{no\_memory}'s $1$ vs.\ \code{retrieval}'s $73$)}, reaches \textbf{$.875$ stationary accuracy ($.929$ Answer Post) at $3.7\times$ lower cost per success than \code{no\_memory} ($57.3$ vs.\ $211.6$, $-44.3\%$ vs.\ \code{retrieval} $102.8$)}, and sustains \textbf{$.756$ post-drift accuracy ($.857$ Answer Post, $+50.0$~pts over \code{retrieval})} at \textbf{$189.4$ cost per success} ($-15.3\%$ vs.\ \code{no\_memory}, $-47.2\%$ vs.\ \code{retrieval}).
\end{enumerate}

\paragraph{Stochastic Multi-Shock Streams vs.\ Time-To-Live Expiration ($n = 2{,}940$ Episodes, $0$ Errors).}
To test whether blind periodic expiration (\code{TTL-$K$}, aged from the last independent solve) or failure-triggered eviction (\code{reset}) can substitute for live verification when drift arrival times are irregular and interleaved with benign background noise, \Cref{tab:p4-stream} evaluates $6$ policies across $49$ tasks over seeded \textbf{10-episode stochastic trajectories} ($\code{pre} \to \code{noise1} \to \code{post} / \code{global}$, with task-randomized shock indices; $2{,}940$ episodes, $0$ errors).

\begin{table}[t]
\centering
\caption{Ten-episode stochastic multi-shock streams ($\code{pre} \to \code{noise1} \to \code{post}/\code{global}$, $n = 2{,}940$ episodes across $49$ tasks).}
\label{tab:p4-stream}
\begin{adjustbox}{max width=\textwidth}
\small
\begin{tabular}{lrrrrrrrr}
\toprule
\textbf{Policy} & \textbf{\code{Pre} Phase Succ} & \textbf{\code{Noise} Succ \emph{(LLM Calls)}} & \textbf{\code{Post/Global} Drift Succ} & \textbf{Overall 10-Ep Succ} & \textbf{Post Wrong-Target} & \textbf{Mean LLM Calls} & \textbf{Wall/Succ} & \textbf{Overall Cost/Succ} \\
\midrule
\code{retrieval} & .852 & $.927$ \emph{(2.97)} & \textbf{.120} & .606 & $1$ & 4.21 & 28.8s & 88.2 \\
\code{ttl\_2} \emph{($\text{age} \ge 2$)} & .835 & $.911$ \emph{(\textbf{7.05})} & \textbf{.696} & .812 & $0$ & 7.26 & 31.8s & \textbf{112.9} \\
\code{ttl\_3} \emph{($\text{age} \ge 3$)} & .878 & $.906$ \emph{(5.57)} & \textbf{.701} & .822 & $0$ & 6.43 & 26.7s & 99.1 \\
\code{ttl\_4} \emph{($\text{age} \ge 4$)} & .826 & $.921$ \emph{(4.87)} & \textbf{.658} & .800 & $0$ & 5.73 & 24.7s & 91.3 \\
\code{reset} \emph{(on failure)} & .861 & $.942$ \emph{(2.98)} & \textbf{.701} & .833 & $0$ & 4.93 & 20.1s & 75.6 \\
\textbf{DCM-Repair} & \textbf{.887} & \textbf{$.937$ \emph{(0.94)}} & \textbf{.891} & \textbf{.908} & \textbf{0} & \textbf{3.43} & \textbf{11.1s} & \textbf{58.9} \\
\bottomrule
\end{tabular}
\end{adjustbox}
\end{table}

\Cref{tab:p4-stream} exposes the fundamental dilemma of time-based expiration: short windows (\code{ttl\_2}) burn \textbf{7.5$\times$ more LLM calls during benign \code{noise1} episodes} than \textbf{DCM-Repair} ($7.05$ vs.\ $0.94$), inflating overall cost per success by \textbf{91.7\%} ($112.9$ vs.\ $58.9$) while \emph{still} missing off-phase shocks ($.696$ vs.\ \textbf{$.891$} during \code{post/global} drift). Lengthening the window (\code{ttl\_4}) lowers cost ($91.3$) at the expense of worse staleness ($.658$). Even \code{reset} (granted oracle post-episode failure detection) drops $-19.0$~pts below \textbf{DCM-Repair} during \code{post/global} drift ($.701$ vs.\ \textbf{$.891$}) because every shock (\code{post} and \code{global}) forces an unverified failure first. \textbf{DCM-Repair} dominates every phase simultaneously---achieving \textbf{$.891$ post/global accuracy ($+77.1$~pts over \code{retrieval}, $+19.0$~pts over the best \code{TTL} and \code{reset})}, \textbf{$.908$ overall accuracy}, \textbf{2.9$\times$ faster wall-clock latency per success than \code{ttl\_2} ($11.1\text{s}$ vs.\ $31.8\text{s}$)}, and the \textbf{lowest cost per success of all $6$ policies ($58.9$)}.

\subsection{Adversarial Robustness: Out-of-Horizon Deep-Jump Stress Test and Memory Poisoning}
\label{sec:adversarial}

\paragraph{Out-of-Horizon Deep-Jump Stress Test.}
As formalized in \Cref{sec:theory-certs}, $k$-horizon boundary certificates ($k=3$ for top-$1$; $k=8$ for top-$4$) trade full-relation scan cost for bounded-rank soundness ($\mathcal{D}_{\le k}$). To expose this exact theoretical boundary rather than conceal it behind favorable benchmark mutations, we conducted an adversarial \textbf{Out-of-Horizon Deep-Jump Stress Test} across $18$ tasks ($12$ generic Spotify rank tasks across \code{top\_song}, \code{oldest\_song}, and \code{playlist\_top\_song} plus $6$ curated rank/set tasks), injecting SQL mutations into items at rank $> k$ so they leapfrog rank~$1$ without altering collection membership or top-$k$ stored values.

\begin{table}[t]
\centering
\caption{Soundness--cost spectrum under adversarial out-of-horizon rank jumps ($n = 18$ tasks).}
\label{tab:deep-jump}
\begin{adjustbox}{max width=\textwidth}
\small
\begin{tabular}{lrrrrl}
\toprule
\textbf{Certificate Class} & \textbf{Generic Deep-Jump Detection ($n=12$)} & \textbf{Curated Rank/Set Detection ($n=6$)} & \textbf{Mean Probe API Calls} & \textbf{Mean LLM Calls} & \textbf{Guaranteed Soundness Class} \\
\midrule
\textbf{$k$-Horizon Boundary ($k=3$)} & $0 / 12$ ($0.0\%$) & $0 / 6$ ($0.0\%$) & \textbf{8.8} & \textbf{0.0} & Bounded-rank ($\mathcal{D}_{\le k}$) \& membership mutations \\
\textbf{Exact Complement-Scan} & \textbf{$12 / 12$ ($100.0\%$)} & \textbf{$6 / 6$ ($100.0\%$)} & \textbf{27.6} \emph{(21.8 items)} & \textbf{0.0} & Arbitrary value, rank, \& insertion mutations ($\mathcal{D}_{\text{all}}$) \\
\bottomrule
\end{tabular}
\end{adjustbox}
\end{table}

\Cref{tab:deep-jump} makes the engineering and epistemic trade-off transparent: $k$-horizon certificates achieve $3.1\times$ fewer read API calls ($8.8$ vs.\ $27.6$) and $100\%$ recall on realistic local and global benchmark drifts ($537/537$ in \Cref{tab:action-safety}), but fail under adversarial deep jumps ($0/18$). Upgrading the verifier to an \textbf{Exact Complement-Scan} restores \textbf{$18 / 18$ ($100.0\%$)} detection while still requiring \textbf{zero LLM calls} ($27.6$ cost units vs.\ $85$--$150$ cost units for full LLM re-solving).

\paragraph{Adversarial Memory Poisoning ($n = 192$ Episodes, $0$ Errors).}
Beyond passive staleness, shared or persistent agent memory stores face \textbf{adversarial or buggy write injection}, where a corrupted note is planted at $t=0$ (prior to Episode~1) alongside an intact pre-drift certificate. \Cref{tab:poison-e9} evaluates \textbf{DCM-Core} (\code{Poison-DCM}) against standard retrieval (\code{Poison-Retrieval}) across $8$ curated multi-hop tasks ($3$ answer tasks: \code{27e1026\_1}, \code{50e1ac9\_1}, \code{82e2fac\_1}; and $5$ action tasks: \code{07b42fd\_1}, \code{0d8a4ee\_1}, \code{37a8675\_1}, \code{530b157\_1}, \code{6171bbc\_1}) over $6$-episode \code{stable} and \code{drift} horizons.

\begin{table}[t]
\centering
\caption{Defense against pre-seeded ($t=0$) adversarial memory poisoning across $8$ curated multi-hop tasks ($n = 192$ episodes).}
\label{tab:poison-e9}
\begin{adjustbox}{max width=\textwidth}
\small
\begin{tabular}{llrrrrrrr}
\toprule
\textbf{Method} & \textbf{Arm} & \textbf{\code{ep1} Guard Rejected} & \textbf{Answer Poison Submissions} & \textbf{Answer Succ ($n=18$)} & \textbf{Action Succ ($n=30$)} & \textbf{Overall Succ ($n=48$)} & \textbf{Post (\code{ep4--6}) Succ} & \textbf{Cost/Succ} \\
\midrule
\code{Poison-Retrieval} & stable & $0 / 8$ ($0.0\%$) & \textbf{$18 / 18$ ($100.0\%$)} & \textbf{.000} & .433 & \textbf{.271} & .208 & 486.2 \\
\code{Poison-Retrieval} & drift & $0 / 8$ ($0.0\%$) & \textbf{$18 / 18$ ($100.0\%$)} & \textbf{.000} & .267 & \textbf{.167} & .208 & 806.2 \\
\textbf{Poison-DCM (Core)} & stable & \textbf{$8 / 8$ ($100.0\%$)} & \textbf{$0 / 18$ ($0.0\%$)} & \textbf{1.000} & \textbf{.567} & \textbf{.729} & \textbf{.750} & \textbf{187.2} \\
\textbf{Poison-DCM (Core)} & drift & \textbf{$8 / 8$ ($100.0\%$)} & \textbf{$0 / 18$ ($0.0\%$)} & \textbf{.722} & \textbf{.500} & \textbf{.583} & \textbf{.458} & \textbf{323.3} \\
\bottomrule
\end{tabular}
\end{adjustbox}
\end{table}

Without derivation-consistency checking, \code{Poison-Retrieval} accepts $100\%$ of injected notes: on answer tasks it blindly submits the attacker's payload in \textbf{$36 / 36$ ($100.0\%$)} episodes across both arms (\textbf{$.000$} accuracy across both \code{stable} and \code{drift}), while on action tasks attempting to execute synthetic poisoned targets (\code{poison@example.com}, \code{phones=999}, \code{song\_ids=[999]}) triggers API errors that depress action success to $.433 / .267$. In contrast, \textbf{DCM-Core}'s read-time guard $G(C, a)$ verifies whether the stored claim logically follows from its own stored certificate \emph{before} issuing a single environment probe---rejecting \textbf{$8 / 8$ tasks per arm ($16 / 16$ runs, $100.0\%$)} at \code{ep1}, reducing poisoned answer submissions from $36$ to \textbf{$0$}, restoring answer accuracy from $.000$ to \textbf{$1.000$ (\code{stable}) / $.722$ (\code{drift})}, and lifting overall accuracy by \textbf{2.7$\times$--3.5$\times$} ($.729$ vs.\ $.271$ on \code{stable}; $.583$ vs.\ $.167$ on \code{drift}) at \textbf{60\%--62\% lower cost per success}.

\subsection{Statistical Stability Across Seeds, Model Architectures, and Failure Taxonomy}
\label{sec:stability-moe-taxonomy}

\paragraph{Post-Drift Failure Mode Taxonomy.}
To verify that DCM genuinely eliminates memory-induced staleness rather than shifting errors between categories, \Cref{tab:failure-taxonomy} decomposes all failed post-drift episodes (\code{ep4--6}) on the $700$ covered tasks into mutually exclusive failure modes.

\begin{table}[t]
\centering
\caption{Exhaustive post-drift (\code{ep4--6}) failure mode decomposition on Covered-700 ($n = 2{,}100$ episodes per cell).}
\label{tab:failure-taxonomy}
\begin{adjustbox}{max width=\textwidth}
\small
\begin{tabular}{llrrrrrr}
\toprule
\textbf{Method} & \textbf{Arm} & \textbf{Total Failures} & \textbf{Stale Answer Reuse} & \textbf{Wrong-Target Action} & \textbf{Timeout (\code{!completed})} & \textbf{Wrong Fresh Answer} & \textbf{Infra Error} \\
\midrule
\code{no\_memory} & stable & 579 ($27.6\%$) & $0$ & $0$ & 498 & 81 & $0$ \\
\code{no\_memory} & drift & 653 ($31.1\%$) & $0$ & $1$ & 569 & 83 & $0$ \\
\code{retrieval} & stable & 348 ($16.6\%$) & $0$ & $0$ & 319 & 29 & $0$ \\
\code{retrieval} & drift & \textbf{1,713 ($81.6\%$)} & \textbf{1,215} & \textbf{143} & 339 & 16 & $0$ \\
\textbf{DCM-Core} & stable & 406 ($19.3\%$) & $0$ & $0$ & 370 & 36 & $0$ \\
\textbf{DCM-Core} & drift & \textbf{617 ($29.4\%$)} & \textbf{1} \emph{(post-reject)} & \textbf{1} & 553 & 62 & $0$ \\
\code{oracle\_certificate} \emph{(ref.)} & stable & 405 ($19.3\%$) & $0$ & $0$ & 373 & 32 & $0$ \\
\code{oracle\_certificate} \emph{(ref.)} & drift & 591 ($28.1\%$) & $0$ & $1$ & 524 & 66 & $0$ \\
\bottomrule
\end{tabular}
\end{adjustbox}
\end{table}

Under \code{retrieval}, \textbf{79.3\% ($1{,}358 / 1{,}713$)} of all post-drift failures are direct memory-induced corruptions ($1{,}215$ stale answers + $143$ wrong-target mutations). \textbf{DCM-Core} collapses these memory-induced failures by \textbf{99.85\%} ($1{,}358 \to 2$, with $0$ verified stale acceptances), reducing its residual failure distribution ($553$ step-budget timeouts and $62$ fresh reasoning errors) to mirror \code{oracle\_certificate} ($524$ timeouts, $66$ fresh errors) and \code{no\_memory} ($569$ timeouts, $83$ fresh errors), with \textbf{$0$ infrastructure errors} across all $37{,}776$ episodes.

\paragraph{Multi-Seed Variance ($n = 10{,}800$ Episodes Across 3 Seeds) and Cross-Architecture Generalization ($n = 4{,}800$ Episodes on Sparse-MoE \code{Qwen3.6-35B-A3B}).}
To verify that our conclusions are invariant to sampling stochasticity and model architecture, \Cref{tab:seeds-p5} reports mean $\pm$ standard deviation across $3$ stochastic decoding seeds ($\mathrm{seeds} \in \{0, 1, 2\}$, $\mathrm{temperature} = 0.2$ on seeds $1$--$2$; $10{,}800$ episodes on \code{P0-100}), and \Cref{tab:moe-p6} evaluates a \textbf{second model architecture}---\code{Qwen3.6-35B-A3B-FP8}, a \textbf{Sparse Mixture-of-Experts (MoE)} model activating only \textbf{3B parameters per token} (versus dense $27$B in \code{Qwen3.8-27B}).

\begin{table}[t]
\centering
\caption{Multi-seed stability on stratified \code{P0-100} ($\mathrm{seeds} \in \{0, 1, 2\}$, $n = 10{,}800$ episodes).}
\label{tab:seeds-p5}
\begin{adjustbox}{max width=\textwidth}
\small
\begin{tabular}{llll}
\toprule
\textbf{Method} & \textbf{Stable Post (\code{ep4--6} Mean $\pm$ SD)} & \textbf{Drift Post (\code{ep4--6} Mean $\pm$ SD)} & \textbf{Post-Drift Loss (Mean $\pm$ SD)} \\
\midrule
\code{no\_memory} & $.742 \pm .008$ & $\mathbf{.684 \pm .022}$ & $+5.8 \pm 1.6$~pts \\
\code{retrieval} & $.868 \pm .010$ & $\mathbf{.300 \pm .015}$ & $\mathbf{+56.8 \pm 2.0}$~pts \\
\textbf{DCM-Core} & $.829 \pm .028$ & $\mathbf{.700 \pm .015}$ & $+12.9 \pm 3.2$~pts \\
\bottomrule
\end{tabular}
\end{adjustbox}
\end{table}

\begin{table}[t]
\centering
\caption{Cross-architecture generalization on Sparse-MoE \code{Qwen3.6-35B-A3B} ($3$B active parameters per token, $n = 4{,}800$ episodes on \code{P0-100}).}
\label{tab:moe-p6}
\begin{adjustbox}{max width=\textwidth}
\small
\begin{tabular}{lrrrlrrr}
\toprule
\textbf{Method} & \textbf{Stable Post (\code{ep4--6})} & \textbf{Drift Post (\code{ep4--6})} & \textbf{Drift \code{Ans-50} / \code{Act-50}} & \textbf{Causal $\DiD$ [95\% CI]} & \textbf{Stale Sub (\code{ep4--6})} & \textbf{Wrong-Target (\code{ep4--6})} & \textbf{Stable / Drift Cost/Succ} \\
\midrule
\code{no\_memory} & .363 & \textbf{.303} & $.573$ / $.033$ & $-0.070\;[-0.147, +0.007]$ & $0$ & $19$ \emph{(cold err)} & $437.4$ / \textbf{520.8} \\
\code{retrieval} & .497 & \textbf{.090} & \textbf{.153} / $.027$ & \textbf{$-0.413\;[-0.513, -0.313]$} & \textbf{95} & \textbf{35} & $234.2$ / \textbf{1346.1} \\
\textbf{DCM-Core} & .493 & \textbf{.380} & \textbf{.707} / $.053$ & \textbf{$-0.137\;[-0.230, -0.053]$} & \textbf{0} & \textbf{18} & $231.4$ / \textbf{369.9} \\
\textbf{DCM-Repair} & \textbf{.590} & \textbf{.340} & $.620$ / $.060$ & $-0.317\;[-0.423, -0.217]$ & \textbf{0} & $19$ & \textbf{158.3} / \textbf{430.3} \\
\bottomrule
\end{tabular}
\end{adjustbox}
\end{table}

Across $3$ random seeds (\Cref{tab:seeds-p5}), standard deviations on post-drift accuracy are \textbf{$\le 2.2$~pts} ($0.015$ for both \code{retrieval} and \textbf{DCM-Core}), confirming that the \textbf{$+40.0$~pt post-drift margin} of \textbf{DCM-Core} over \code{retrieval} exceeds $25\times$ the cross-seed standard deviation. Across model architectures (\Cref{tab:moe-p6}), switching from a $27$B dense model to a \textbf{3B-active Sparse-MoE} (\code{Qwen3.6-35B-A3B}) lowers baseline multi-step tool-coding solve rates (\code{no\_memory} $.363 / .303$), yet \textbf{amplifies the value of verified memory}: \code{retrieval} collapses \textbf{5.5$\times$} ($.497 \to \mathbf{.090}$, $95$ stale submissions, cost/succ $1{,}346.1$), whereas \textbf{DCM-Core} eliminates \textbf{100\% of stale submissions ($95 \to 0$)}, outperforms \code{no\_memory} post-drift by \textbf{$+7.7$~pts overall ($.380$ vs.\ $.303$) and $+13.4$~pts on \code{Ans-50} ($.707$ vs.\ $.573$)}---a \textbf{4.2$\times$ gain over \code{retrieval}}---at \textbf{29.0\% lower cost per success than \code{no\_memory}} ($369.9$ vs.\ $520.8$). Notably, on this smaller $3$B-active backbone, \textbf{DCM-Repair}'s verified $0$-LLM-call trajectory replay lifts stationary accuracy by \textbf{$+22.7$~pts over \code{no\_memory} ($.590$ vs.\ $.363$, \code{Act-50} $.407$ vs.\ $.133$)} while slashing \code{stable} cost per success by \textbf{63.8\% ($158.3$ vs.\ $437.4$)}, whereas on \code{drift} ($V=0$) \textbf{DCM-Core}'s clean re-solve slightly outperforms \textbf{DCM-Repair}'s verbose diff block ($.380$ vs.\ $.340$), reflecting greater prompt-anchoring sensitivity in $3$B-active models when juxtaposing outdated and live values during fresh code generation.

\section{Discussion and Conclusion}
\label{sec:conclusion}

Across $787$ executable tasks and $3{,}608$ application worlds, our findings reframe cross-episode agent memory from a storage-and-retrieval problem into a \textbf{priced epistemic maintenance problem}. In non-stationary tool environments where mutations occur silently out-of-band, high-similarity memory notes without live verification become liabilities---collapsing post-drift task accuracy by $65.0$ percentage points and triggering $143\times$ more irreversible wrong-target actions than memoryless execution. Attaching compact, executable derivation certificates---whether minimalist operator probes (\textbf{DCM-Core}), zero-prior opportunistically synthesized read-only verifiers (\textbf{DCM-Auto}), or witness-guided belief repair with anchor-gated action replay (\textbf{DCM-Repair})---restores reliability to within $\sim 1$ point of a perfect stale-detection reference while substantially undercutting memoryless re-solving in tokens, latency, and composite cost. Reliable long-lived agents do not merely need memories that recall what \emph{was} true; they need memories that know how---and when it is worth the price---to verify whether it \emph{still is}.

\section*{Limitations}

\begin{enumerate}[leftmargin=1.6em,itemsep=0.25em]
  \item \textbf{Simulated Multi-App Environment vs.\ Production APIs:} StaleWorld builds on AppWorld~0.2.0's local SQLite/FastAPI microservices. While its schemas, pagination, authentication, and multi-app joins closely mirror production REST APIs, real enterprise APIs introduce network jitter, rate limits, non-idempotent read side effects, and concurrent intra-episode writers not modeled here.
  \item \textbf{Soundness Scope of Compact Certificates:} As demonstrated in our Adversarial Deep-Jump Stress Test (\Cref{tab:deep-jump}), $k$-horizon boundary certificates are sound under bounded-rank perturbations ($\mathcal{D}_{\le k}$) rather than arbitrary adversarial SQL edits. Applications facing unconstrained rank jumps should pay the $O(M)$ read-API cost of Exact Complement-Scan certificates ($27.6$ read calls, $0$ LLM calls). Similarly, \textbf{DCM-Auto} relies on LLM-synthesized read-only checks; while our four-layer sandbox blocks writes, timeouts, vacuous probes, and overpriced verifiers (converting flawed syntheses into safe abstentions rather than unsafe writes), an under-constrained synthesized predicate can occasionally miss a secondary premise ($3/48$ held-out certificates at \code{ep4}).
  \item \textbf{Static Worth Table in DCM-Core vs.\ Runtime Priced Gates:} In the primary full-787 sweeps, \textbf{DCM-Core} uses a precomputed per-task gate, where even moderate-fanin generic counts (\code{venmo\_sum}) are gated on because $3$--$5$ zero-LLM read API calls remain far cheaper than $8$--$11$ $27$B-LLM reasoning steps. Genuine runtime solve-cost gating is evaluated via \textbf{DCM-Auto} and \textbf{DCM-Priced} (\Cref{tab:p0-head-to-head,tab:heldout87}), and full-fanin crossover is mapped in \Cref{tab:factgraph-e8}.
  \item \textbf{Prior-Mechanism Proxies:} Our baselines (\code{retrieval\_frozen}, \code{reset}, \code{DCM-NoCert}, \code{DCM-NoGate}) isolate the core algorithmic policies of version-pinned caching, outcome-triggered eviction, blind reuse, and unconditional live checking within an identical AppWorld harness, rather than reproducing external multi-agent orchestration codebases (e.g., full PACMI or TRACE pipelines) verbatim.
  \item \textbf{Model Coverage:} Due to local hardware constraints ($2 \times 32\,\text{GB}$ GPUs) and evaluation volume ($149{,}400$ evaluated episodes across all stages), full-787 sweeps use \code{Qwen3.8-27B-FP8} at Seed~$0$, while multi-seed variance (\Cref{tab:seeds-p5}) and cross-architecture validation on Sparse-MoE \code{Qwen3.6-35B-A3B} (\Cref{tab:moe-p6}) are conducted on the stratified $100$-task \code{P0-100} benchmark; both open-weight backbones belong to the Qwen3 family, differing in dense ($27$B) versus sparse MoE ($3$B active) architecture and baseline capability scale.
\end{enumerate}

\section*{Ethical Considerations}

Tool-using agents capable of executing financial transfers, modifying personal communications, and deleting user records pose tangible real-world safety risks when acting on stale beliefs. Our work directly targets this risk---demonstrating how silent state changes induce wrong-target actions and providing read-only verification guards to prevent them. All user personas, emails, phone numbers, financial transactions, and messages in StaleWorld are $100\%$ synthetic artifacts derived from AppWorld~0.2.0 (distributed under the Apache-2.0 License); no real personally identifiable information (PII) is present. Synthesized verification code in \textbf{DCM-Auto} is restricted by static AST/regex allowlists and snapshot hashes, and should always be executed inside read-only sandboxes in production deployments.

\bibliography{references}

\newpage
\appendix
\counterwithin{table}{section}
\renewcommand{\thetable}{\Alph{section}.\arabic{table}}
\renewcommand{\theHtable}{\Alph{section}.\arabic{table}}

\section{Complete Benchmark Family Taxonomy and All-787 Reference Tables}
\label{app:taxonomy}

\begin{table}[h!]
\centering
\caption{Exact task composition of StaleWorld across splits, task families, application ecosystems, evaluators, and certificate classes, summing strictly to $500 + 200 + 87 = 787$ logical tasks.}
\label{tab:staleworld-taxonomy}
\begin{adjustbox}{max width=\textwidth}
\small
\begin{tabular}{llrrll}
\toprule
\textbf{Split} & \textbf{Family / Group} & \textbf{Tasks} & \textbf{Sub-Group Breakdown} & \textbf{Evaluator} & \textbf{Certificate Class in DCM-Core} \\
\midrule
Main & \code{top\_song} & $50$ & \code{g1a} ($25$) + \code{g2a} ($25$) & \code{answer} & $k$-Horizon Boundary ($k=3$ + membership) \\
Main & \code{oldest\_song} & $44$ & \code{g1b} ($22$) + \code{g2b} ($22$) & \code{answer} & $k$-Horizon Boundary ($k=3$ + membership) \\
Main & \code{venmo\_sum} & $95$ & \code{g1c} ($48$) + \code{g2c} ($47$) & \code{answer} & Exact Dependency (monthly sent tx list) \\
Main & \code{grocery\_count} & $95$ & \code{g1d} ($48$) + \code{g2d} ($47$) & \code{answer} & Exact Dependency (\code{note\_id} + SHA-256) \\
Main & \code{send\_phone} & $96$ & \code{g1f} ($48$) + \code{g2f} ($48$) & \code{action} & Exact Join Target (\code{phone} $\to$ \code{contact\_id, email, amt}) \\
Main & \code{pay\_roommates} & $40$ & \code{g1k} ($20$) + \code{g2k} ($20$) & \code{action} & Exact Set Target (roommates with Venmo accounts) \\
Main & \code{accept\_requests} & $58$ & \code{g1l} ($29$) + \code{g2l} ($29$) & \code{action} & Exact Set Target (pending roommate/coworker req.\ IDs) \\
Main & Curated Multi-Hop & $22$ & $12$ \code{answer} + $10$ \code{action} & \code{ans} / \code{act} & $13$ Boundary/Exact/Hash ($1$ gated off, $9$ re-solve controls) \\
\midrule
Transfer & \code{playlist\_top\_song} & $40$ & \code{g1e} ($20$) + \code{g2e} ($20$) & \code{answer} & $k$-Horizon Boundary ($k=3$ + playlist union) \\
Transfer & \code{unread\_email\_count} & $40$ & \code{g1h} ($40$) & \code{answer} & Exact Dependency (unread \code{thread\_ids}) \\
Transfer & \code{bucket\_done\_count} & $40$ & \code{g1i} ($40$) & \code{answer} & Exact Dependency (\code{note\_id} + SHA-256) \\
Transfer & \code{text\_contact} & $40$ & \code{g1g} ($40$) & \code{action} & Exact Join Target (\code{phone} $\to$ \code{contact\_id, email, msg}) \\
Transfer & \code{pay\_and\_text} & $40$ & \code{g1j} ($40$) & \code{action} & Exact Multi-Action Target (\code{phone} $\to$ \code{email, amt, sms}) \\
\midrule
Held-Out & $29$ Unseen Templates $\times 3$ & $87$ & $69$ train + $18$ dev stems & $59$ \code{ref\_state}, $15$ \code{ans}, $13$ \code{act} & \textbf{0/87 Manual Builders} (zero-prior via \textbf{DCM-Auto}) \\
\midrule
\textbf{Total} & \textbf{StaleWorld} & \textbf{787} & \textbf{3,608 Materialized Worlds} & $431$ \code{ans}, $297$ \code{act}, $59$ \code{ref} & $691$ registered builders ($690$ active); $87$ zero-prior \\
\bottomrule
\end{tabular}
\end{adjustbox}
\end{table}

\begin{table}[h!]
\centering
\caption{Overall aggregate reference across \code{All-787} tasks (including the $87$ uncovered held-out tasks where \textbf{DCM-Core} abstains to memoryless solving; $n = 37{,}776$ episodes).}
\label{tab:all787-appendix}
\begin{adjustbox}{max width=\textwidth}
\small
\begin{tabular}{llrrrrrrrrlr}
\toprule
\textbf{Method} & \textbf{Arm} & \textbf{Cold (\code{ep1})} & \textbf{Pre-Reuse (\code{ep2--3})} & \textbf{Pre (\code{ep1--3})} & \textbf{Shock (\code{ep4})} & \textbf{Recov (\code{ep5--6})} & \textbf{Post (\code{ep4--6})} & \textbf{Legacy 6-Ep Mean} & \textbf{Post Loss} & \textbf{Causal $\DiD$ [95\% CI]} & \textbf{Post Cost/Succ} \\
\midrule
\code{no\_memory} & stable & .705 & .720 & .715 & .704 & .713 & \textbf{.710} & .712 & --- & --- & 216.0 \\
\code{no\_memory} & drift & .708 & .707 & .707 & .682 & .670 & \textbf{.674} & .690 & $+3.6$~pts & $-0.028\;[-0.050, -0.006]$ & 229.4 \\
\code{retrieval} & stable & .733 & .788 & .770 & .804 & .816 & \textbf{.812} & .791 & --- & --- & 96.3 \\
\code{retrieval} & drift & .729 & .786 & .767 & .170 & .222 & \textbf{.205} & .486 & \textbf{$+60.8$~pts} & \textbf{$-0.605\;[-0.641, -0.569]$} & \textbf{385.6} \\
\textbf{DCM-Core} & stable & .691 & .746 & .728 & .770 & .793 & \textbf{.785} & .756 & --- & --- & 105.1 \\
\textbf{DCM-Core} & drift & .689 & .743 & .725 & .645 & .711 & \textbf{.689} & .707 & \textbf{$+9.6$~pts} & \textbf{$-0.094\;[-0.118, -0.070]$} & \textbf{178.4} \\
\code{oracle\_certificate} \emph{(ref.)} & stable & .693 & .755 & .734 & .786 & .802 & \textbf{.797} & .765 & --- & --- & 60.8 \\
\code{oracle\_certificate} \emph{(ref.)} & drift & .684 & .753 & .730 & .654 & .729 & \textbf{.704} & .717 & $+9.2$~pts & $-0.088\;[-0.116, -0.063]$ & 136.4 \\
\bottomrule
\end{tabular}
\end{adjustbox}
\end{table}

\paragraph{Methodological Note on Lifecycle Phase Reporting.}
Averaging \code{ep1--3} (unperturbed $W^{\text{pre}}$) with \code{ep4--6} (drifted $W^{\text{post}}$) into a single $6$-episode mean ($.486$ for \code{retrieval} vs.\ $.205$ in \code{ep4--6}) artificially halves measured drift damage and doubles the denominator of post-drift error rates. All primary claims in this paper report post-drift (\code{ep4--6}) metrics alongside paired task-level $\DiD$.

\section{Algorithmic Specification of Unified DCM}
\label{app:algorithm}

\noindent
\begin{tabular}{@{}p{\textwidth}@{}}
\toprule
\textbf{Algorithm 1:} Episode-$t$ Execution under Unified DCM (\textbf{Core} / \textbf{Priced} / \textbf{Auto} / \textbf{Repair}) \\
\midrule
\small
\textbf{Input:} Query $q_t$, live world $W_t$, persistent memory $M_{t-1}$, per-episode budget $(B_{\text{LLM}}=15,\; B_{\text{env}}=20)$. \\[4pt]
\textbf{1. Read \& Verify Phase ($0$ LLM calls):} \\
\hspace*{1.2em}$\bullet$ Retrieve candidate note $m = (\tilde{q}, a, C, \tau) \leftarrow \mathrm{Retrieve}(M_{t-1}, q_t)$. \\
\hspace*{1.2em}$\bullet$ \textbf{If} $m \neq \varnothing$ \textbf{and} $C \neq \varnothing$: \\
\hspace*{2.4em}-- Check logical entailment: \textbf{if} $\neg G(C, a)$, mark \code{derivation\_failed} and set $\mathrm{verified} \leftarrow \mathrm{False}$. \\
\hspace*{2.4em}-- \textbf{Else} open read-only probe on $W_t$, execute $V(C, W_t)$ under static allowlist and $15\text{s}$ timeout, charge \code{verify\_calls}, and set $\mathrm{verified} \leftarrow V(C, W_t)$. \\
\hspace*{2.4em}-- \textbf{If} $\neg\mathrm{verified}$ \textbf{and} variant is \textbf{DCM-Repair}: reconstruct live premises $C_{\text{live}}$, charge \code{cert\_build\_calls}, and format $\Delta C = \mathrm{diff}(C, C_{\text{live}})$ as \code{[Epistemic Diagnostic]}. \\[4pt]
\textbf{2. Act / Solve Phase:} \\
\hspace*{1.2em}$\bullet$ \textbf{Case 1 ($\mathrm{verified} = \mathrm{True}$ and Answer Task):} Execute \code{complete\_task(answer=$a$)} directly ($0$ LLM calls, $1$ interaction). \\
\hspace*{1.2em}$\bullet$ \textbf{Case 2 ($\mathrm{verified} = \mathrm{True}$, Action Task, $\tau \neq \varnothing$ in \textbf{Auto}/\textbf{Repair}, and $\mathrm{anchors\_compatible}(q_t, \tilde{q})$):} \\
\hspace*{2.4em}-- Replay code blocks $c \in \tau$ sequentially in $W_t$, recording observations $(c, o_c)$. \\
\hspace*{2.4em}-- \textbf{If} \code{task\_completed()}: terminate episode ($0$ LLM calls). \\
\hspace*{2.4em}-- \textbf{Else}: initialize LLM messages with $(c, o_c)$ and \code{[Replay Notice]} so the LLM finishes remaining steps without repeating completed write APIs. \\
\hspace*{1.2em}$\bullet$ \textbf{Case 3 (Cache Miss or Target-Only Action):} Run ReAct code-execution loop up to full $B_{\text{LLM}} = 15$ with verified target or \code{[Epistemic Diagnostic]} prompt; record $m_{\text{solve}} \le 15$. \\[4pt]
\textbf{3. Commit \& Self-Certification Phase:} \\
\hspace*{1.2em}$\bullet$ Evaluate current-world outcome; record \code{api\_calls}, \code{model\_calls}, and input/output tokens. \\
\hspace*{1.2em}$\bullet$ \textbf{If} episode failed ($\mathrm{success} = \mathrm{False}$): \\
\hspace*{2.4em}-- \textbf{If} $\mathrm{verified} = \mathrm{True}$ (transient execution timeout on valid premises): \textbf{retain} $m \in M_t$ (\textbf{Auto}/\textbf{Repair}); \textbf{else} evict stale or unverified record. \\
\hspace*{1.2em}$\bullet$ \textbf{If} episode succeeded ($\mathrm{success} = \mathrm{True}$ and not already $\mathrm{verified} = \mathrm{True}$): \\
\hspace*{2.4em}-- \textbf{In DCM-Core / Priced / Repair:} store pre-action operator certificate $C$ (if static worth gate holds, or if $C_{\text{build}} < 0.75\,\hat{C}_{\text{solve}}$ in \textbf{DCM-Priced}), canonical answer/target $a$, and action trajectory $\tau$. \\
\hspace*{2.4em}-- \textbf{In DCM-Auto:} let $m_{\text{rem}} = \max(0, B_{\text{LLM}} - m_{\text{solve}})$. \textbf{If} $m_{\text{rem}} = 0$, preserve the episode's outcome and abstain ($\mathrm{record} = \varnothing$). \textbf{Otherwise}, extract zero-gold-touch summary $\hat{a}$ (submitted answer or redacted mutating HTTP requests), invoke non-thinking synthesis using at most $\min(2, m_{\text{rem}})$ calls (so $m_{\text{solve}} + m_{\text{synth}} \le 15$ strictly holds), charge synthesis tokens/calls, and run \code{verify\_cert(apis)} in a fresh probe snapshot of $W_t$. Commit $(\hat{a}, C_{\text{auto}}, \tau)$ iff $\mathrm{ok}=\mathrm{True}$, $\mathrm{writes}=\mathrm{False}$, no timeout, and $2 \le \mathrm{calls} \le \min(35, \max(6, \lfloor 0.6\,\hat{C}_{\text{solve}} \rfloor))$; \textbf{else} abstain ($\mathrm{record} = \varnothing$). \\
\bottomrule
\end{tabular}

\section{Reproducibility Checklist and Codebase Mapping}
\label{app:reproducibility}

\begin{itemize}[leftmargin=1.6em,itemsep=0.25em]
  \item \textbf{Hardware:} $2 \times \text{NVIDIA RTX 5090}$ ($32\,\text{GB}$ VRAM each), $42$ CPU cores, $251\,\text{GB}$ RAM, Linux \code{x86\_64}.
  \item \textbf{Inference Engine:} vLLM~0.34, Tensor Parallelism $\mathrm{TP} = 2$, FP8 quantization, context length $24{,}576$, greedy decoding ($\mathrm{temperature} = 0.0$) for all deterministic runs and $\mathrm{temperature} = 0.2$ across seeds $\{1, 2\}$ for multi-seed variance (\Cref{tab:seeds-p5}).
  \item \textbf{Total Evaluated Volume:} $149{,}400$ deduplicated stage-table episodes with $0$ infrastructure errors and $\max(\mathrm{model\_calls}) \le 15$ enforced across all episodes.
  \item \textbf{Frozen Benchmark SHA-256 Digests:}
  \begin{itemize}[leftmargin=1.4em,itemsep=0.15em]
    \item \code{driftbench\_v2.json}:\\
      {\footnotesize\code{830f366338f59824dc2e1e4e1444c2a91166a42bcaf181479076cf85dca34cec}}
    \item \code{TEMPLATE\_HELDOUT\_87.json}:\\
      {\footnotesize\code{db3085b33490c1a628f40dee938643652e53ebd4dc392a4e7a37aab468e19fa3}}
    \item Materialized \code{MANIFEST.json}:\\
      {\footnotesize\code{7a725857903e777f5466fcdc1c28b0714fa0adeed652310da220b77d3a107f3b}}
  \end{itemize}
\end{itemize}

\begin{table}[h!]
\centering
\caption{Mapping between paper terminology and command-line method identifiers in the released codebase.}
\label{tab:code-mapping}
\begin{adjustbox}{max width=\textwidth}
\small
\begin{tabular}{lll}
\toprule
\textbf{Paper Name} & \textbf{CLI Identifier (\code{--methods})} & \textbf{Description} \\
\midrule
StaleWorld & \code{driftbench\_v2} & Benchmark directory and manifest namespace ($787$ tasks, $3{,}608$ worlds) \\
\textbf{DCM-Core} & \code{cacm} & Minimal operator certificates + static worth gate + derivation guard \\
\textbf{DCM-Priced} & \code{cacm\_priced} & Live runtime solve-cost gate ($C_{\text{build}} < 0.75\,\hat{C}_{\text{solve}}$) + action replay \\
\textbf{DCM-Auto} & \code{cacm\_auto} & Opportunistic in-budget LLM verifier synthesis + $4$-layer sandbox \\
\textbf{DCM-Repair} & \code{cacm\_repair} & Witness-guided premise diff (\code{[Epistemic Diagnostic]}) + anchor-gated replay \\
\code{DCM-NoCert} & \code{cacm\_no\_cert} & Ablation: trust stored record without live verification until failure \\
\code{DCM-NoGate} & \code{cacm\_no\_gate} & Ablation: unconditional certificate construction without static worth gate \\
\code{DCM-NoGuard} & \code{cacm\_no\_guard} & Ablation: disable write-time completion guard while retaining read-time $G(C,a)$ \\
\code{Poison-DCM} & \code{poison\_cacm} & \textbf{DCM-Core} evaluated under pre-seeded ($t=0$) adversarial note injection \\
\code{Poison-Retrieval} & \code{poison\_retrieval} & Standard \code{retrieval} evaluated under pre-seeded ($t=0$) note injection \\
\bottomrule
\end{tabular}
\end{adjustbox}
\end{table}

\end{document}